\documentclass[10pt,twocolumn,letterpaper]{article}

\usepackage{iccv}
\usepackage{times}
\usepackage{epsfig}
\usepackage{graphicx}
\usepackage{amsmath}
\usepackage{amssymb}
\usepackage{adjustbox}

\usepackage{times}
\usepackage{epsfig}
\usepackage{graphicx}
\usepackage{amsmath}
\usepackage{amssymb}

\usepackage{color}
\usepackage[table]{xcolor}
\usepackage{multirow}
\usepackage{tabularx}
\usepackage{hhline}
\usepackage[para,online,flushleft]{threeparttable}
\usepackage{stackengine,scalerel}

\definecolor{red}{rgb}{0.95,0.4,0.4}
\definecolor{blue}{rgb}{0.4,0.4,0.95}
\definecolor{darkblue}{rgb}{0,0,0.8}
\definecolor{darkred}{rgb}{0.8,0,0}
\definecolor{darkgreen}{rgb}{0.15,0.6,0.15}
\definecolor{grey}{rgb}{0.6,0.6,0.6}
\definecolor{col1}{RGB}{232, 161, 148}
\definecolor{col2}{RGB}{148, 187, 232}

\newcommand\overstar[1]{\ThisStyle{\ensurestackMath{%
  \setbox0=\hbox{$\SavedStyle#1$}%
  \stackengine{0mm}{\copy0}{\kern.2\ht0\smash{\SavedStyle*}}{O}{c}{F}{T}{S}}}}

\usepackage[pagebackref=true,breaklinks=true,letterpaper=true,colorlinks,bookmarks=false]{hyperref}

\iccvfinalcopy 

\def\iccvPaperID{6092} 

\ificcvfinal\fi

\begin{document}

\title{Self-supervised Monocular Depth Estimation: Let's Talk About The Weather}

\author{Kieran Saunders\\
Aston University\\
Birmingham, UK\\
{\tt\small 190229315@aston.ac.uk}
\and
George Vogiatzis\\
Loughborough University\\
Leicestershire, UK\\
{\tt\small g.vogiatzis@lboro.ac.uk}
\and
Luis J. Manso\\
Aston University\\
Birmingham, UK\\
{\tt\small l.manso@aston.ac.uk}
}

\maketitle
\ificcvfinal\thispagestyle{empty}\fi

\begin{abstract}
    Current, self-supervised depth estimation architectures rely on clear and sunny weather scenes to train deep neural networks. However, in many locations, this assumption is too strong. For example in the UK (2021), 149 days consisted of rain. For these architectures to be effective in real-world applications, we must create models that can generalise to all weather conditions, times of the day and image qualities. Using a combination of computer graphics and generative models, one can augment existing sunny-weather data in a variety of ways that simulate adverse weather effects. While it is tempting to use such data augmentations for self-supervised depth, in the past this was shown to degrade performance instead of improving it. In this paper, we put forward a method that uses augmentations to remedy this problem. By exploiting the correspondence between unaugmented and augmented data we introduce a pseudo-supervised loss for both depth and pose estimation. This brings back some of the benefits of supervised learning while still not requiring any labels. We also make a series of practical recommendations which collectively offer a reliable, efficient framework for weather-related augmentation of self-supervised depth from monocular video. We present extensive testing to show that our method, Robust-Depth, achieves SotA performance on the KITTI dataset while significantly surpassing SotA on challenging, adverse condition data such as DrivingStereo, Foggy CityScape and NuScenes-Night. The project website can be found \href{https://kieran514.github.io/Robust-Depth-Project/}{here}.
\end{abstract}

\section{Introduction}
\label{sec:intro}

Depth estimation has been a pillar of computer vision for decades and has many applications, such as self-driving, robotics, and scene reconstruction. While multiple-view geometry is a well-understood computer vision problem, the advent of deep learning has enabled depth estimation from a single image. The first such methods used a supervised approach to estimate depth and required expensive ground truth data collected using LIDAR and Radar sensors. Recently, self-supervised monocular methods have been introduced, using photometric loss \cite{zhou2017unsupervised} to achieve view synthesis on consecutive images as a form of self-supervision. These methods have received wide interest because of their low cost and ability to generalize to multiple scenarios \cite{hu2020seasondepth}. 
However, despite its potential, self-supervised Monocular Depth Estimation (MDE) has been hampered by adverse weather conditions and nighttime driving ~\cite{wang2021regularizing,vankadari2022sun}.
\begin{figure}[t]
  \centering
   \includegraphics[width=1\linewidth]{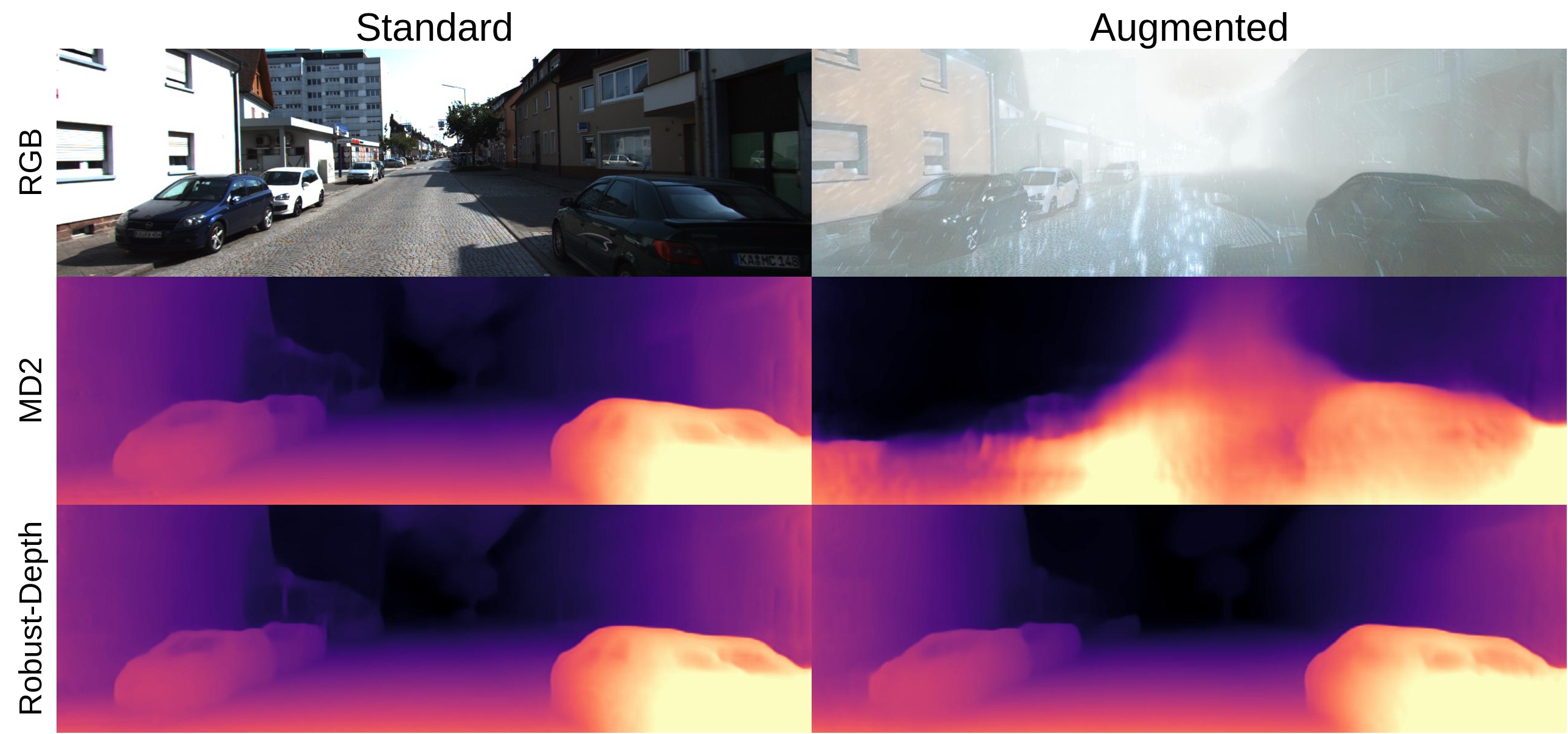}
   \caption{Monodepth2 demonstrates impressive performance on sunny scenes but struggles in different weather scenarios. Our method is more robust to environmental changes.}
   \label{fig:onecol}
    \vspace{-0.4cm}
\end{figure}
The KITTI dataset \cite{geiger2012we}, used extensively for MDE, only contains daytime, dry, sunny images rather than varied realistic conditions.
Previous attempts have shown that training these methods on different domains, like nighttime data, lead to worse performance \cite{liu2021self}.
Figure \ref{fig:onecol} illustrates Monodepth2 \cite{godard2019digging}, trained on the KITTI dataset, estimating depth accurately on sunny data but struggles when shifting domains to different weather conditions. Our method \textbf{Robust-Depth}, trained with the KITTI dataset and augmentations, is more robust to such changes. 


Unlike in other fields of computer vision \cite{zhong2020random, perez2017effectiveness}, dataset augmentation has led to worse performance for MDE (see Table \ref{ablation}).
One possible explanation is that augmentations lead to texture shifts, and we know that CNNs are poor at generalising to texture shifts \cite{bae2023study}.
It is known \cite{dijk2019neural,peng2021excavating,wang2021regularizing} that depth networks become reliant on vertical cues (e.g., the output for a pixel being dependent on its vertical position in an image). Pixels at the bottom of the image are assumed to be close and pixels at the top are assumed to be far from the camera. While \cite{wang2021regularizing} suggests this is beneficial overall, these methods rely on potentially false cues. For example, a cliff-side image would pose difficulties to a system over-relying on vertical cues. In this work, we put forward a further observation that has to do with self-supervision itself: When performing data augmentation in supervised learning, we introduce some form of noise on the input image while maintaining a noise-free label. On the other hand, in self-supervised methods, the labels come from the data themselves. So augmenting a data point in a self-supervised method introduces noise in both data and target labels, leading to a much harder machine learning problem. 

To overcome this challenge we propose a different formulation of the self-supervised loss, which exploits the alignment between the unaugmented and augmented data. Under that scheme, we are able to treat depth maps obtained from the unaugmented data as soft targets for the augmented depth estimations. Furthermore, with minimal extra effort, we can also treat depth maps of augmented images as labels for the unaugmented depth predictions leading to a fully symmetric bi-directional consistency constraint. We call this \textit{pseudo-supervision loss} because it is an attempt to leverage the benefits of supervised learning (faster learning rates) in the self-supervised domain. For more detail see section \ref{Bi-directional Proxy Depth Supervision Hints}.
The paper also puts forward a number of recommendations for creating a robust, data-augmentation framework for MDE, each of which helps overcome the reliance on simplistic cues for depth. These include:
\begin{itemize}
    \item Using the unaugmented images when warping the target image with the current depth map (sec. \ref{Bi-directional Proxy Depth Supervision Hints} eq. \eqref{semi-aug})
    \item Training in (unaugmented, augmented) pairs (sec. \ref{Bi-directional Proxy Depth Supervision Hints} eq. \eqref{optimse_both})
    \item Applying a one-way pseudo-supervision loss for pose estimation (sec. \ref{Bi-directional Proxy Depth Supervision Hints} eq. \eqref{posers})
\end{itemize}
Finally, we propose a wide-ranging set of weather-related data augmentations together with vertical cropping and tilling, the effect of which is to move the focus of the network away from simplistic depth cues and towards deeper semantic cues. The paper contains an extensive experimental analysis, including an ablation study of the various algorithmic components as well as a comparison to State-of-the-Art (SotA). The analysis shows that our method significantly surpasses SotA on adverse weather data and performs as well or better on sunny weather data.

\section{Related Work}
\subsection{Supervised Depth}
Initially, depth estimation was cast as a simple regression task using ground truth depth data collected via one or multiple sensors. CNNs were first used for MDE in \cite{eigen2014depth}, and there have been many architectural and network developments over time \cite{9316778,lee2019big,agarwal2022attention}. Furthermore, supervised methods made the regression task a classification-regression task \cite{fu2018deep,bhat2021adabins,li2022binsformer} as the regression tasks lead to substandard results. These networks relied on high-cost ground truth data and struggled with generalizing to other datasets.

\subsection{Self-supervised Depth}
\subsubsection{Stereo}
Garg \textit{et al.} \cite{garg2016unsupervised} used view synthesis as a method for self-supervised learning for stereo depth between stereo pairs. Later, Monodepth \cite{godard2017unsupervised} used photometric loss, made from $L_1$ loss and SSIM \cite{wang2004image}, to demand left-right consistency between left-right reconstruction.

\subsubsection{Monocular}
Our work focuses on monocular cameras, as stereo setups are more expensive and require more space. The first technique to use view synthesis for MDE was SfM-learner \cite{zhou2017unsupervised}.
This and the methods to follow use a depth network with pose estimations to warp images and maximize photometric uniformity. Monodepth2 (MD2) \cite{godard2019digging} introduced the per-pixel minimization of the photometric error to avoid occlusion issues while also introducing auto-masking to handle texture-less regions and dynamic objects. 


One issue with MDE is that pose, and therefore depth, can only be estimated up to an unknown scale causing Monodepth2 to struggle with maintaining consistent scales between frames. To counter this, SC-SfM-Learner \cite{bian2021unsupervised} created a differentiable geometric loss and masked out pixels with geometric inconsistency. This led to higher inter-frame depth consistency but is computationally more expensive for each iteration.
Other approaches make use of cost volumes for their efficiency, specifically, Manydepth~\cite{watson2021temporal} and MonoRec~\cite{wimbauer2021monorec} use temporal image sequences that allow for geometrical reasoning during inference. They use multiple pre-defined depths hypotheses to warp reference frame features to the target image frame, and their differences create a cost volume. The depth hypothesis closest to the true depth leads to low values in the cost volume. 

Moreover, RA-depth \cite{he2022ra} developed a useful augmentation that leads to more scale consistency and better depth estimation when changing resolutions. This is crucial as these self-supervised methods have no reference for scale and, at the very least, should have consistent depth in the same image at different scales. 



In parallel to new training methodologies, there is work on improving architectures, and some recent research has been focused on increasing efficiency while increasing accuracy \cite{lyu2021hr,guizilini20203d,zhou2021self,lite-mono}. We will focus on applying our contributions to DepthNet from Monodepth2~\cite{godard2019digging} and the current SotA depth network \cite{zhao2022monovit} to show improvements. As demonstrated in \cite{varma2022transformers, bae2023study} transformer networks are more robust than their CNN equivalent and self-supervised methods are more robust and generalisable than supervised methods. 
An explanation for this is that transformer networks are more shape-biased versus the texture bias of CNNs, allowing transformers to be more accurate on out-of-distribution data and texture variations (such as style transfers).
This inspires us to use MonoViT~\cite{zhao2022monovit} as our final model because it is lightweight, accurate and more robust than a pure CNN network. We must note however that the methodology proposed in this paper is agnostic to the specifics of the neural model used for depth or pose estimation.

\subsection{Data Augmentation}
Many previous methods have attempted to solve the lack of robustness problem with MDE \cite{wang2021regularizing,vankadari2022sun,spencer2020defeat,gurram2021monocular,atapour2018real,liu2021self}. A common theme with these papers is that they can only handle few variations in environments and require large and complex architectural modifications to achieve realistic depth for the augmentations. Some also heavily rely on synthetic data, which leads to the models having a domain bias (from real to synthetic) \cite{atapour2018real,gurram2021monocular}. To our knowledge, there has been a limited use of augmentations in MDE \cite{ishii2021cutdepth, kim2022global, semi_sup}, especially self-supervised MDE. Simple colour jitters and horizontal flips have been used in the past but, more complicated augmentations have been avoided as they are typically found to degrade depth accuracy.

ADDS-DepthNet \cite{liu2021self} is the most similar to our method as they create night image pairs from day images using GANs. In doing so, they train their network to be effective in two domains, night and day. They use the day depth estimation as pseudo-supervision for their depth estimation for night scenes. This method is one-directional and limits the depth estimation to only be as good as the day depth estimation. Also, they ignore pose estimation and focus on creating a reconstruction of the GAN-generated night image. We avoid reconstruction of the augmented images as the GAN-generated augmentations lack consistency between frames, especially for the night domain.
Furthermore, our model is unique in that it is able to handle multiple domains with greater accuracy.




Another approach,  ITDFA \cite{zhao2022unsupervised}, uses a trained fixed depth decoder from Monodepth2 \cite{godard2019digging} and changes the encoders for each domain to force consistency between features in different domains. Also, ITDFA enforces consistency between depth images for both day and night. This has obvious drawbacks compared to our model; firstly, we use the same encoder and decoder for all different augmentations, which can be trained end-to-end. Secondly, the depth estimations from these augmented images are limited to the accuracy of Monodepth2. Our method avoids this issue by only updating the augmented depth where the unaugmented depth has a lower photometric loss (sec. \ref{Bi-directional Proxy Depth Supervision Hints}).  

Furthermore, \cite{dijk2019neural} highlighted the fact that MDE tends to learn only a small number of cues and shows great dependency on the position of the pixel to estimate depth. EPC-depth~\cite{peng2021excavating} demonstrated that the use of data grafting for input images leads to greater performance with respect to stereo depth. Inspired by this and to help MDE learn new and more varied cues, we apply vertical cropping to the input images. Also, we apply tile cropping to the input images, to enforce equivariance for depth estimation. Similarly, RA-depth \cite{he2022ra} used augmentation to change the scales of the input image, allowing the depth estimation to be more scale consistent. A modified version of this idea has been employed in our method.


\section{Method}\label{Method}
\begin{figure}[h]
  \centering
   \includegraphics[width=1\linewidth]{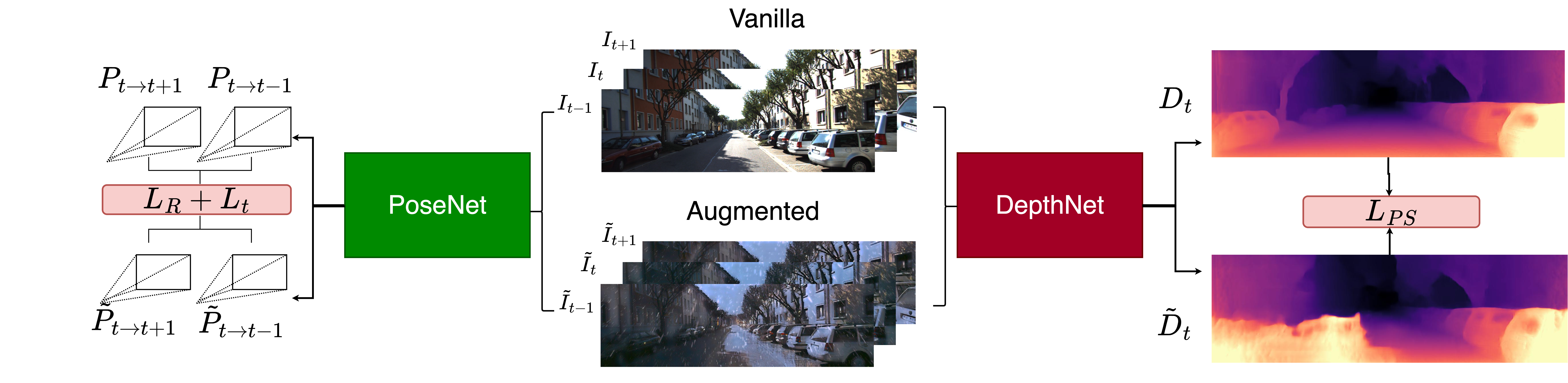}
   \caption{The original image (top-middle) and the augmented image (bottom-middle) are each fed into the pose and depth estimation networks. Our pseudo-supervision loss function contains terms that encourage consistency between the depth map and pose estimates of the original and augmented image.}
   \label{fig:new}
    \vspace{-0.4cm}
\end{figure}

\textbf{Overview:}
In this paper, we put forward \textbf{Robust-Depth}, a self-supervised architecture that is able to estimate depth and ego-motion while being robust to many different augmentations. We highlight SotA's limited robustness in other weather scenarios, over-dependence on pixel position, and its lack of ability to generalise to other datasets. All of this is achieved with a minor increase in computation and memory requirements while using the standard DepthNet backbone from Monodepth2 \cite{godard2019digging} and the current SotA network MonoViT \cite{zhao2022monovit}. 
The applications of our novel contributions are shown in Figure \ref{fig:new}. 


\subsection{Preliminaries}
Following Zhou et al. \cite{zhou2017unsupervised} we simultaneously train an ego-motion network with a depth network to execute view-synthesis between consecutive frames. We use the target depth estimation $D_{t}=DepthNet(I_t)$ and camera pose estimations $P_{t \to t^\prime} = PoseNet(I_t,I_t^\prime)$ to synthesise the target image where $I_{t^\prime} \in \{I_{t-1},I_{t+1}\}$, only using source frames. PoseNet and DepthNet denote the pose and depth estimation networks respectively. The synthesised image obtained is projected using inverse warping as shown below.
\begin{equation} \label{eq1}
    I_{{t^\prime} \to t} = I_{t^\prime} \langle Proj(D_{t}, P_{t \to {t^\prime}}, K) \rangle	
\end{equation}
Proj() outputs the resulting 2D coordinates of the depths after projecting into the camera of frame $I_{t^\prime}$, and $\langle \rangle$ is the sampling operator. We will be using photometric loss (pe) which is defined below,
\begin{equation} \label{pe}
    pe(I_a,I_b)=\frac{\alpha}{2}(1-SSIM(I_a,I_b))+(1-\alpha)||I_a-I_b||,
\end{equation}
where $\alpha=0.85$. In Monodepth2, the per-pixel photometric loss used for training the pose and depth network is simply:
\begin{equation} \label{min}
    L_p = \min_{t^\prime}pe(I_t, I_{t^\prime \to t})
\end{equation}

Where for each pixel we decide whether to use the next or the previous frames to reproject, based on the minimization of reprojection error. 
Moreover, unlike some methods that advocate using only one layer of the depth network to increase learning speed~\cite{lee2021learning,he2022ra}, we use the multi-scale proposal from Monodepth2.




\subsection{Bi-directional pseudo-supervision Loss} \label{Bi-directional Proxy Depth Supervision Hints}

Previous methods that used augmented images for self-supervised MDE attempted to create reconstructions of the augmented images \cite{liu2021self}, as shown below. 
\begin{equation}\label{augs}
    \Tilde{I}_{{t^\prime} \to t} = \Tilde{I}_{t^\prime} \langle Proj(\Tilde{D}_t, \Tilde{P}_{t \to {t^\prime}}, K) \rangle
\end{equation}
Where $\Tilde{I}_{t^\prime}$ represents the augmented input image, $\Tilde{D}_t$ is the depth estimation of the augmented image, and $\Tilde{P}_{t \to {t^\prime}}$ is the pose estimation obtained from the augmented images. They would then proceed with the standard loss function; 
\begin{equation} \label{min}
    L_p = \min_{t^\prime}pe(\Tilde{I}_t, \Tilde{I}_{t^\prime \to t})
\end{equation}

The augmented input images are generally created using GANs and lack consistency between consecutive frames. This will cause the reconstructed augmented images $\Tilde{I}_{{t^\prime} \to t}$ to be dissimilar to the true target image as there may be changes between frames that can not be accounted for, e.g. illumination changes and rain streaks. Furthermore, the pose estimation will suffer from the same problem as it requires consecutive frame inputs, which in the case of augmented images, will not be consistent. 
To counter these issues we introduce semi-augmented warping;
\begin{equation}\label{semi-aug}
    \overstar{I}_{{t^\prime} \to t} = I_{t^\prime} \langle Proj(\Tilde{D}_t, P_{t \to {t^\prime}}, K) \rangle.
\end{equation}
Semi-augmented warping exploits the consistency between the unaugmented frames while using the depth estimations from the augmented images. Furthermore, we use the pose estimation from the unaugmented input to ensure that the depth estimation is isolated from the errors of pose estimation under the various augmentations.

Simply optimising the photometric loss for the augmented reprojected image from equation \eqref{augs} can lead to catastrophic forgetting. This happens because the network shifts its focus onto the augmented data and forgets the original data (see Table \ref{ablation}). To prevent this we always train images in pairs (augmented, unaugmented) as follows.
\begin{equation} \label{optimse_both}
    L_p = \min_{t^\prime}(pe(I_t, I_{{t^\prime} \to t})) + \min_{t^\prime}(pe(I_t, \overstar{I}_{{t^\prime} \to t})).
\end{equation}
There are several possible formulations of the loss function that achieve the same objective, however, we found this simple pair-training scheme works well in practice. 



The photometric loss given above is responsible for enforcing the self-supervision constraint. However, the relationship between the augmented and unaugmented data gives us an opportunity to exploit the faster learning rate of supervised learning. To see this, assume we had access to a fully trained network that can correctly predict depth in the unaugmented, original images. 
We note that each augmented image should correspond to the same underlying depth map as the original image it was derived from since all augmentations considered do not change the underlying depth map. We could therefore obtain that depth map using the fully trained network on the original image, and then use that depth map as a label in a supervised regression loss.

In practice, we do not have access to such a fully trained network on the original images. During training, we can expect that our depth network will either perform better in the augmented or the unaugmented image at any given time. However, using the photometric reprojection loss, we can easily find out which of the two depths gives a better reprojection and use that depth as a label for the other depth map. Furthermore, this analysis can be done per-pixel leading to an efficient scheme which we term pseudo-supervision loss.

We define the following per-pixel mask that picks out pixels where the unaugmented image gives a better depth than the augmented one
\begin{equation} \label{5}
    M_{v} = [\min_{t^\prime}pe(I_t, I_{t^\prime \to t}) < \min_{t^\prime}pe(I_t, \overstar{I}_{t^\prime \to t})] \odot \mu,
\end{equation}
where [ ] are the Iverson brackets and $\odot$ is the pixel-wise product. We define $\mu$ as auto-masking from Monodepth2 \cite{godard2019digging}, which is applied to our mask $M_{v}$ to remove stationary pixels.
We apply this same concept in the opposite direction for pixels where the augmented depth leads to a smaller reprojection loss than the unaugmented depth estimation.
\begin{equation} \label{6}
    M_{a} = [\min_{t^\prime}pe(I_t, \overstar{I}_{t^\prime \to t}) < \min_{t^\prime}pe(I_t, I_{t^\prime \to t})] \odot \mu
\end{equation}
Note, we cannot simply find the inverse of this mask, as we have to account for the auto-masked pixels $M_a \neq 1 - M_v$.
We can now introduce the components of bi-directional pseudo-supervision depth loss;
\begin{equation}
    L_{a} = log(|\Tilde{D}_{t} - \underline{{D}_{t}}| + 1) \odot M_{v},
\end{equation}
\begin{equation}
    L_v = log(|D_{t} - \underline{\Tilde{D}_{t}}| + 1) \odot M_a.
\end{equation}
The underlining of $\underline{D_{t}}$ and $\underline{\Tilde{D}_t}$ signifies that the gradients of $D_{t}$ and $\Tilde{D}_{t}$ are cut off during backpropagation.
This gives us the bi-directional pseudo-supervision loss:
\begin{equation}
    L_{PS} = L_{a} + L_v.
\end{equation}
$L_{a}$ is the loss caused by a worse augmented depth estimation compared to the unaugmented depth estimation, and $L_v$ is the loss caused by a worse unaugmented depth estimation than augmented depth estimation, summing to make the bi-direction pseudo-supervision loss $L_{PS}$. We show the contribution of each component to the bi-direction pseudo-supervision loss in the supplementary materials. 

\begin{table*}[h!]
    \centering
    \begin{adjustbox}{max width=\textwidth}
    \footnotesize
    \begin{tabular}{|l|c|c|c|c|c|l||c|c|c|c|c|c|c|} \hline
        \textbf{Ablation} & 
        
        \begin{tabular}{@{}c@{}}Eq. \eqref{optimse_both}\end{tabular} & 
        \begin{tabular}{@{}c@{}}Eq. \eqref{semi-aug} \\ Depth\end{tabular} &
        \begin{tabular}{@{}c@{}}Eq. \eqref{semi-aug} \\ Pose\end{tabular} & 
        \begin{tabular}{@{}c@{}}$L_{PS}$\end{tabular} &
        \begin{tabular}{@{}c@{}}$L_R$+ \\ $L_t$\end{tabular} &

        \textbf{Test} & \cellcolor{col1}Abs Rel & \cellcolor{col1}Sq Rel & \cellcolor{col1}RMSE 
        & \cellcolor{col1}RMSE log & \cellcolor{col2}$\delta < 1.25 $ & \cellcolor{col2}$\delta < 1.25^{2}$ & \cellcolor{col2}$\delta < 1.25^{3}$ \\ \hline
        
        \multirow{2}{*}{Monodepth2 (\textit{Sunny}) \cite{godard2019digging}} & & & & & & \textit{Sunny} &
        {0.115} & {0.903} & {4.863} &   {0.193} &   {0.877} &   {0.959} &   {0.981} \\ 
        &  & & & & & \cellcolor{lightgray} \textit{Bad w.} &  \cellcolor{lightgray} 0.249  &  \cellcolor{lightgray} 2.477  &  \cellcolor{lightgray} 7.962  & \cellcolor{lightgray}  0.347  & \cellcolor{lightgray}  0.612  &  \cellcolor{lightgray} 0.824  &  \cellcolor{lightgray} 0.918  \\ \hline

        \multirow{2}{*}{Monodepth2 (\textit{Bad w.}) \cite{godard2019digging}} & & & & & & \textit{Sunny} &   0.118  &   0.920  &   4.814  &   0.193  &   0.869  &   0.958  &   0.981  \\
        & & & & & & \cellcolor{lightgray} \textit{Bad w.} &  \cellcolor{lightgray} 0.140  &  \cellcolor{lightgray} 1.069  &   \cellcolor{lightgray}5.408  &  \cellcolor{lightgray} 0.220  & \cellcolor{lightgray}  0.821  & \cellcolor{lightgray}  0.940  & \cellcolor{lightgray}  0.975  \\ \hline \hline

        \multirow{2}{*}{Robust-Depth} & \multirow{2}{*}{\checkmark} & & & & & \textit{Sunny} &   0.116  &   0.948  &   4.884  &   0.194  &   0.875  &   0.959  &   0.981  \\
        & & & & & & \cellcolor{lightgray} \textit{Bad w.} &  \cellcolor{lightgray} 0.155  &  \cellcolor{lightgray} 1.623  & \cellcolor{lightgray}  5.677  &  \cellcolor{lightgray} 0.232  & \cellcolor{lightgray}  0.812  & \cellcolor{lightgray}  0.933  &  \cellcolor{lightgray} 0.970  \\ \hline

        \multirow{2}{*}{Robust-Depth} & \multirow{2}{*}{\checkmark} & \multirow{2}{*}{\checkmark} &  & & & \textit{Sunny} &   0.116  &   0.881  &   4.912  &   0.196  &   0.871  &   0.958  &   0.980  \\
        & & & & & & \cellcolor{lightgray} \textit{Bad w.} & \cellcolor{lightgray}  0.135  &  \cellcolor{lightgray} 1.048  & \cellcolor{lightgray}  5.351  &  \cellcolor{lightgray} 0.216  &  \cellcolor{lightgray} 0.830  &  \cellcolor{lightgray} 0.944  &  \cellcolor{lightgray} 0.976  \\ \hline

        \multirow{2}{*}{Robust-Depth} & \multirow{2}{*}{\checkmark} & & \multirow{2}{*}{\checkmark} & & & \textit{Sunny} &   0.116  &   0.890  &   4.872  &   0.192  &   0.869  &   0.959  &   0.982  \\
        & & & & & & \cellcolor{lightgray} \textit{Bad w.} &  \cellcolor{lightgray} 0.138  &  \cellcolor{lightgray} 1.066  &  \cellcolor{lightgray} 5.481  &  \cellcolor{lightgray} 0.219  &  \cellcolor{lightgray} 0.821  &  \cellcolor{lightgray} 0.941  & \cellcolor{lightgray}  0.976  \\ \hline

        \multirow{2}{*}{Robust-Depth} & \multirow{2}{*}{\checkmark} & & & \multirow{2}{*}{\checkmark} & & \textit{Sunny} &   0.116  &   0.931  &   4.816  &   0.193  &   0.877  &   0.959  &   0.981  \\
        & & & & & & \cellcolor{lightgray} \textit{Bad w.} &  \cellcolor{lightgray} 0.139  &   \cellcolor{lightgray} 1.136  & \cellcolor{lightgray}  5.439  &  \cellcolor{lightgray} 0.219  &  \cellcolor{lightgray} 0.827  &  \cellcolor{lightgray} 0.942  & \cellcolor{lightgray}  0.975  \\ \hline

        \multirow{2}{*}{Robust-Depth} & \multirow{2}{*}{\checkmark} & & & & \multirow{2}{*}{\checkmark} & \textit{Sunny} &   0.116  &   0.942  &   4.895  &   0.194  &   0.874  &   0.958  &   0.981  \\
        & & & & & & \cellcolor{lightgray} \textit{Bad w.} & \cellcolor{lightgray}  0.146  &  \cellcolor{lightgray} 1.249  &  \cellcolor{lightgray} 5.533  &   \cellcolor{lightgray} 0.226  & \cellcolor{lightgray}  0.818  & \cellcolor{lightgray}  0.937  &  \cellcolor{lightgray} 0.972  \\ \hline

        \multirow{2}{*}{Robust-Depth All} & \multirow{2}{*}{\checkmark} & \multirow{2}{*}{\checkmark} & \multirow{2}{*}{\checkmark} & \multirow{2}{*}{\checkmark} & \multirow{2}{*}{\checkmark} & \textit{Sunny} &   0.115  &   0.937  &   4.840  &   0.193  &   0.873  &   0.959  &   0.981  \\
        & & & & & &  \cellcolor{lightgray} \textit{Bad w.} &  \cellcolor{lightgray} 0.133  &  \cellcolor{lightgray} 1.115  & \cellcolor{lightgray}  5.259  &  \cellcolor{lightgray} 0.211  & \cellcolor{lightgray}  0.842  & \cellcolor{lightgray}\cellcolor{lightgray}  0.948  & \cellcolor{lightgray}  0.977  \\ \hline
            
    \end{tabular}
    \end{adjustbox}
    \vspace{1mm}
    \caption{\textbf{Ablation Study:} We demonstrate the addition of each component to the baseline Monodepth2. Monodepth2 (\textit{Sunny}) represents Monodepth2 trained on the original KITTI dataset and (\textit{Bad w.}) represents training with augmentations. Each method is tested in the \textit{sunny} and \textit{bad weather} datasets (white and grey rows respectively).}
    
    \label{ablation}
\end{table*}
The analysis above also applies to the pose estimation part of the algorithm. The rotation/translation that links two consecutive frames remains the same after these images are augmented. Once again, we can introduce a bi-directional consistency loss using reprojection quality to judge which pose to use as a label. However in experiments, this was found to have negligible improvement, so we assume that the pose estimated from the original images will be closer to the correct one. This leads to the pseudo-supervised pose loss, 
\begin{equation}\label{posers}
    L_{R} = |\Tilde{R}-\underline{R}|, \quad L_{t} = |\Tilde{t}-\underline{t}|,
\end{equation}
where $\Tilde{R}$ and $\Tilde{t}$ are the rotation and translation obtained from the augmented images and $\underline{R}$, $\underline{t}$ from the original. The underlining again denotes that the gradients of $R$ and $t$ are not used when updating the pose network weights.

\textbf{Final Loss:}
Our final loss is a combination of photometric loss, bi-directional pseudo-supervision loss, pseudo-supervision pose loss, and edge-aware smoothness loss ($L_s$) from \cite{ranjan2019competitive}. Where edge-aware smoothness loss is applied to both the unaugmented depth and augmented depth. Giving a final loss, which averages over each pixel, scale and batch;
\begin{equation}
    L = \mu L_p + \omega L_{PS} + \beta (L_R + L_t) + \gamma L_s.
\end{equation}
\subsection{Data preparation \& augmentations}\label{data}
Previous depth estimation methods struggled with variations in weather. Fog, rain, extreme brightness, night time and motion blur have all been a challenge for all such depth networks. There have been few attempts at solving some of these issues, but they suffer from poor results \cite{wang2021regularizing,vankadari2022sun,spencer2020defeat,zhao2022unsupervised}, or they use simulation data that struggles with the domain shift to real-world application \cite{atapour2018real,gurram2021monocular}. 

We use a physics-based renderer (PBR) \cite{tremblay2021rain} to create realistic rain and fog augmentations on the KITTI dataset and store these prior to training the presented architecture. Similarly, we use CoMoGAN \cite{pizzati2021comogan} to augment the KITTI dataset with realistic night-time, dawn and dusk. Furthermore, we make all the combinations of these weather-based augmentations, for example, rain+fog or rain+fog+dawn. Separately, we also add ground snow as it is a great source of corruption and motion blur using Automold \cite{Automold}.
We use grey-scale images and break the image down into its red, green and blue components to remove the effects of colour on depth estimation. This is important for large shifts in image colour, for example, between autumn and winter.
To achieve extreme brightness, we increase the brightness capabilities on the already implemented colour jitter.

To further improve the robustness of self-supervised MDE, we took inspiration from robustness testing on other tasks \cite{michaelis2019benchmarking,hendrycks2019robustness}. \cite{varma2022transformers} demonstrated the robustness of monocular depth using transformer depth networks with corruption augmentations on the test sets. Rather than solely relying on the underlying network for robustness, we implement these augmentations prior to training. Specifically, we add Gaussian noise, shot noise, impulse noise, defocus blur, frosted glass blur, zoom blur, snow, frost, elastic, pixelated, and jpeg augmentation variants to the KITTI dataset. The percentage and distribution of the augmentations are provided in the supplementary materials.

During training, we use a modified scale augmentation from RA-Depth \cite{he2022ra}, vertical cropping augmentation, and tile augmentation created especially for this architecture so that the augmentations can be reverted later. Furthermore, standard random erase is used as an augmentation. Also, note that due to the lack of variety of data used for self-supervised monocular depth, all of these augmentations will reduce the chance that MDE overfits naive cues such as vertical dependency, texture bias and apparent size of objects.  

\textbf{Vertical cropping augmentation:}
This augmentation was inspired by EPC-Depth \cite{peng2021excavating} grafting augmentation. We randomly select value $\tau$ between \{0.2,0.4,0.6,0.8\} with uniform probability. We then select this percent of the input image from the top down and crop it to be moved underneath the section below it, all of this occurs on each minibatch. 

\textbf{Tile augmentation:}
For tile cropping, we split the width of the image by two or four, and the height of the image by two or three randomly. 
We crop each section and then randomly shuffle the tiles to a new position, all of this occurs on each minibatch.
\section{Results}
\textbf{Experimental set-up:}
We train the model starting with pretrained ImageNet weights \cite{deng2009imagenet} using PyTorch \cite{NEURIPS2019_9015} and train on an NVIDIA A6000 GPU. We use the Adam optimiser \cite{kingma2014adam} for 30 epochs, with an input size of $640\times 192$ and set a starting learning rate of $10^{-4}$. Progressively reducing the learning rate using multi-step learning rate decay at epoch $[20,25,29]$ by 0.1. We set the hyperparameters $\omega$, $\beta$ and $\gamma$ to 0.01, 0.01 and 0.001, respectively. 
In all tables, Robust-Depth$^\dagger$ represents Robust-Depth trained with just weather and time-specific augmentations. 
Finally, Robust-Depth$^*$ represents Robust-Depth using MonoViTs transformer backbone instead of the baseline (Monodepth2) ResNet18.
\subsection{Datasets}
\textbf{KITTI} \cite{geiger2012we}:
We use the official Zhou split for validation data consisting of 4,424 images and train on the entire 39,810 images from the training set. For testing, we use 697 images from the Eigen \textit{et al.} test set \cite{eigen2015predicting}. 
An in-depth discussion of how we created the variants of this dataset is provided in the supplementary material. This includes all the generation parameters that determine how to reproduce the augmentations when replicating the experiments. All methods results to follow are trained on the KITTI dataset, unless specified. By \textit{sunny} we denote the raw, unaugmented KITTI dataset while \textit{bad weather} denotes the KITTI dataset augmented with adverse weather effects and other image modifications as laid out in sec. \ref{data}.

\textbf{DrivingStereo} \cite{yang2019drivingstereo}:
As the aim of this paper is to demonstrate the capacity to infer depth in different domains, the DrivingStereo dataset provides a selection of foggy, cloudy, rainy and sunny images for testing, all containing 500 images. 

\textbf{Foggy Cityscape} \cite{SDV18, cordts2016cityscapes}:
To test methods on more severe foggy scenes, we use the Foggy CityScape dataset as we use the data provided with the beta parameter equal to 0.02 (most severe fog) consisting of 1525 test images.

\textbf{NuScenes} \cite{caesar2020nuscenes}:
Nighttime scenarios are a challenging domain for many self-supervised monocular depth methods, and we can use NuScenes to explicitly test the ability of each method to infer depth in this domain.  We select the nighttime splits from \cite{wang2021regularizing} test selection and test on 500 real nighttime images. 
\begin{table*}[t]
  \centering
  \tiny
  \resizebox{1.0\textwidth}{!}{
    \begin{tabular}{|l|c|l||c|c|c|c|c|c|c|}
        \arrayrulecolor{black}\hline
          Method & W$\times$H & Test & \cellcolor{col1}Abs Rel & \cellcolor{col1}Sq Rel & \cellcolor{col1}RMSE  & \cellcolor{col1}RMSE log & \cellcolor{col2}$\delta < 1.25 $ & \cellcolor{col2}$\delta < 1.25^{2}$ & \cellcolor{col2}$\delta < 1.25^{3}$ \\
         
        \hline\hline
                
        \multirow{2}{*}{Ranjan \etal \cite{ranjan2019competitive}} & \multirow{2}{*}{832$\times$256} & \textit{Sunny} & 0.148 & 1.149 & 5.464 & 0.226 & 0.815 & 0.935 & 0.973 \\ 
        & & \cellcolor{lightgray} \textit{Bad w.} & \cellcolor{lightgray}0.262 & \cellcolor{lightgray}2.263 & \cellcolor{lightgray}7.757 & \cellcolor{lightgray}0.3560 & \cellcolor{lightgray}0.570 & \cellcolor{lightgray}0.811 & \cellcolor{lightgray}0.919 \\
        \hline

        \multirow{2}{*}{Monodepth2 \cite{godard2019digging}} & \multirow{2}{*}{640$\times$192} & \textit{Sunny} &
         {0.115} &   {0.903} &   {4.863} &   {0.193} &   {0.877} &   {0.959} &   {0.981} \\ 
         & & \cellcolor{lightgray} \textit{Bad w.} & \cellcolor{lightgray} 0.249  &  \cellcolor{lightgray} 2.477  & \cellcolor{lightgray}  7.962  & \cellcolor{lightgray}  0.347  &  \cellcolor{lightgray} 0.612  &  \cellcolor{lightgray} 0.824  & \cellcolor{lightgray}  0.918  \\
         \hline

        \multirow{2}{*}{Mono-Uncertainty\cite{Poggi_CVPR_2020}} & \multirow{2}{*}{640$\times$192} & \textit{Sunny} & 0.111 & 0.863 & 4.756 & 0.188 & 0.881 & 0.961 & 0.982\\
        && \cellcolor{lightgray} \textit{Bad w.} & \cellcolor{lightgray}  0.243  & \cellcolor{lightgray}  2.287  &  \cellcolor{lightgray} 7.806  & \cellcolor{lightgray}  0.341  &  \cellcolor{lightgray} 0.613  &  \cellcolor{lightgray} 0.831  &  \cellcolor{lightgray} 0.923  \\
        \hline

        \multirow{2}{*}{HR-Depth~\cite{lyu2021hr}} & \multirow{2}{*}{640$\times$192} & \textit{Sunny} & 0.109 & 0.792 & 4.632 & 0.185 & 0.884 & 0.962 & 0.983\\
        & & \cellcolor{lightgray} \textit{Bad w.} &  \cellcolor{lightgray} 0.273  &  \cellcolor{lightgray} 2.544  & \cellcolor{lightgray}  8.408  &  \cellcolor{lightgray} 0.380  &  \cellcolor{lightgray} 0.552  &  \cellcolor{lightgray} 0.793  &  \cellcolor{lightgray} 0.901  \\
        \hline

        \multirow{2}{*}{CADepth~\cite{yan2021channel}} & \multirow{2}{*}{640$\times$192} & \textit{Sunny} & 0.105 & 0.769 & 4.535 & 0.181 & 0.892 & 0.964 & 0.983\\ 
        & &\cellcolor{lightgray}  \textit{Bad w.} &  \cellcolor{lightgray} 0.266  & \cellcolor{lightgray}  2.530  &  \cellcolor{lightgray} 8.145  &  \cellcolor{lightgray} 0.365  &  \cellcolor{lightgray} 0.575  &  \cellcolor{lightgray} 0.801  &  \cellcolor{lightgray} 0.908  \\
        \hline

        \multirow{2}{*}{DIFFNet ~\cite{zhou2021self}} & \multirow{2}{*}{640$\times$192} & \textit{Sunny} & 0.102 & 0.749 & \underline{4.445} & 0.179 & \underline{0.897} & 0.965 & 0.983\\
        & &\cellcolor{lightgray} \textit{Bad w.} &  \cellcolor{lightgray} 0.202  & \cellcolor{lightgray}  1.724  & \cellcolor{lightgray}  7.198  & \cellcolor{lightgray}  0.304  &  \cellcolor{lightgray} 0.680  &  \cellcolor{lightgray} 0.870  &  \cellcolor{lightgray} 0.940  \\
        \hline
        
        %

        \multirow{2}{*}{MonoViT~\cite{zhao2022monovit}} & \multirow{2}{*}{640$\times$192} & \textit{Sunny} & \bf{0.099} & \bf{0.708} & \bf{4.372} & \bf{0.175} & \bf{0.900} & \bf{0.967} & \bf{0.984}\\
        & & \cellcolor{lightgray} \textit{Bad w.} &  \cellcolor{lightgray} 0.167  & \cellcolor{lightgray}  1.347  & \cellcolor{lightgray}  6.385  &  \cellcolor{lightgray} 0.258  & \cellcolor{lightgray}  0.751  &  \cellcolor{lightgray} 0.909  &  \cellcolor{lightgray} 0.961  \\
        \hline

        \hline
        \multirow{2}{*}{\textbf{Robust-Depth}} & \multirow{2}{*}{640$\times$192} & \textit{Sunny} &   0.115  &   0.937  &   4.840  &   0.193  &   0.873  &   0.959  &   0.981  \\
        & & \cellcolor{lightgray} \textit{Bad w.} &  \cellcolor{lightgray} \underline{0.133}  & \cellcolor{lightgray}  \underline{1.115}  & \cellcolor{lightgray}  \underline{5.259}  & \cellcolor{lightgray}  \underline{0.211}  & \cellcolor{lightgray}  \underline{0.842}  &   \cellcolor{lightgray}\underline{0.948}  &  \cellcolor{lightgray} \underline{0.977}  \\
        \hline

        \multirow{2}{*}{\textbf{Robust-Depth$^*$}} & \multirow{2}{*}{640$\times$192} & \textit{Sunny} &   \underline{0.100}  &   \underline{0.747}  &   4.455  &   \underline{0.177}  &   0.895  &   \underline{0.966}  &   \textbf{0.984}  \\
        & & \cellcolor{lightgray} \textit{Bad w.} & \cellcolor{lightgray}  \textbf{0.114}  &  \cellcolor{lightgray} \textbf{0.891}  & \cellcolor{lightgray}  \textbf{4.878}  &  \cellcolor{lightgray} \textbf{0.193}  &  \cellcolor{lightgray} \textbf{0.868}  &  \cellcolor{lightgray} \textbf{0.958}  &  \cellcolor{lightgray} \textbf{0.981}  \\
        
        \arrayrulecolor{black}\hline

    \end{tabular}}  
  \vspace{1mm}
  \caption{\textbf{Quantitative Results for the KITTI Eigen Test Dataset.} Note that we do not use any test time refinement and only require one frame to do inference for depth estimation. All methods have been pretrained on ImageNet \cite{deng2009imagenet} and \textbf{bold} text represents the best result for the metric for each test. Whereas, the \underline{underlined} text represents the second best for each metric for each test.
  \label{tab:kitti_eigen}} 
\vspace{-5mm}
\end{table*}
\subsection{Ablation Study:}
This subsection provides a detailed discussion of the results of the ablation study (Table \ref{ablation}). We test using the \textit{sunny} and \textit{bad weather} datasets to demonstrate quantitatively how well a model can handle sunny weather as well as image degradation and varying weather conditions, including time of day.
Initially, when testing Monodepth2 with the \textit{bad weather} test set (row 1), we obtain a disappointing performance, indicating a lack of ability to handle weather conditions and image degradation. 

\textbf{Optimising for augmented data (Row 2):}
When attempting to train Monodepth2 with augmented data, in the naive way we outline in equations \eqref{augs} and \eqref{min}, we see a significant improvement in robustness but worse performance for the \textit{sunny} test. Here we train with augmented data in the traditional sense, including unaugmented images. These results inspired our method, which by design, attempts to maintain a balance between focus on sunny and weather-augmented data.

\textbf{Optimising for unaugmented \& augmented data (Row 3):}
Optimising for both the \textit{sunny} and \textit{bad weather} data simultaneously (see equation \eqref{optimse_both}) leads to the best of both worlds, a form of trade-off. Here we emphasise that we are using the augmented reprojection from equation \eqref{augs} and not semi-augmented warping from equation \eqref{semi-aug}. This formulation achieves \textit{sunny} test results almost as good as the standard Monodepth2 and yet only slightly worse performance for robustness compared to naively training Monodepth2 with augmented data. The worse performance for the \textit{bad weather} testing is because the network trained on only augmented data sufferers from catastrophic forgetting. All rows to follow will optimise for both \textit{sunny} and \textit{bad weather} data.

\textbf{Semi-augmented Warping (Row 4 \& 5):}
We explore the effect of equation \eqref{semi-aug} into two different parts. 
Firstly, we investigate the effects of warping the original image while using the pose and depth obtained from the augmented image (row 4). From this, we see that, due to the inconsistency of the augmented images, there is a significant improvement from exploiting the consistency of unaugmented images. 
Secondly, we evaluate the effect of using the unaugmented pose estimation while warping the augmented images (row 5). Again, a notable reduction is shown in \textit{bad weather} test metrics. This affirms our assumption that the unaugmented pose estimations will result in the most accurate pose, as the pose network relies heavily on consistency between two frames.



\textbf{Pseudo-supervision depth loss (Row 6):}
We aim for consistency between unaugmented and augmented depth estimations. This also leads to significant loss reductions in the \textit{bad weather} test set, which supports the use of bi-directional pseudo-supervision, using unaugmented image depth estimations to improve the depth estimations of augmented images and vice versa.

\textbf{Pseudo-supervision pose loss (Row 7):}
Although using unaugmented pose estimation for inverse warping augmented images is beneficial, we can still try to improve the robustness of the pose network by using the pseudo-supervised scheme for improving pose estimation. This again shows a reduction in \textit{bad weather} metrics, and we would expect this is making the pose network more accurate. The reasoning behind this is that the GANs inconsistency will cause the pose of the augmented images to be worse than the unaugmented pose, and the pose loss will help the pose network be more robust.

\textbf{All (Row 8):}
Our model trained with all of the contributions leads to \textit{sunny} test results close to Monodepth2 and the best \textit{bad weather} performance.
We use Monodepth2 as our baseline, but we expect to see the same pattern when using any network. 
The pseudo-supervised depth loss leads to a 10.3\% reduction. On the other hand, semi-augmented warping (equation \eqref{semi-aug}) leads to a 12.9\% reduction in the absolute relative error for the \textit{bad weather} test.
Therefore, out of all the algorithmic components, warping with unaugmented images leads to the most significant reduction in the \textit{bad weather} test metrics. In the final row, we confirm that our method of training with augmented data leads to the best performance on the augmented data, significantly outperforming the baseline model trained with augmented data, while maintaining \textit{sunny} data depth performance. 

\subsection{Comparison with SotA}

Baseline models in Table \ref{tab:kitti_eigen} have not been trained on augmented data, but, just like we demonstrated in Table \ref{ablation} when applying our contribution to any of these methods as a baseline we would expect to see the same patterns; improved \textit{bad weather} metrics and maintained \textit{sunny} metrics. 

Table \ref{tab:kitti_eigen} shows our model's results on the KITTI Eigen test set \cite{eigen2015predicting} compared with previous and current state-of-the-art methods/architectures for self-supervised MDE. There is strong evidence that Robust-Depth is able to perform at the same level as Monodepth2 for the original KITTI dataset, but performs significantly better than all previous state-of-the-art for the \textit{bad weather} test set. This demonstrates the depth networks' robustness to a variety of image degradation, weather and time of day changes. 

Note that the previous state-of-the-art, along with improved methods, have mainly developed more sophisticated architectures which have led to improvements in depth. Many recent works have demonstrated better generalisability to domain changes from transformer networks.
To make this comparison fair we have trained our model with the standard Monodepth2 (ResNet18) architecture and a second model with the same transformer backbone as MonoViT (Robust-Depth$^*$). 
From Robust-Depth$^*$, we see maintained \textit{sunny} depth abilities and impressive \textit{bad weather} results, further demonstrating that our method is able to handle augmentations and thus changes in domains. 


\subsection{Further Testing}
The data used to do the robustness tests up to this point are sampled from the same distribution that our model was trained on. To truly demonstrate our model's capabilities on all of these weather changes and nighttime scenes, we test on out-of-distribution data.
\definecolor{aqua}{rgb}{0.0, 1.0, 1.0}
\begin{table}[t!]
\centering
\resizebox{\columnwidth}{!}{%
\renewcommand{\tabcolsep}{1mm}
\begin{tabular}{|c|l||c|c|c|c|c|c|c|}
\hline
Domain & Method & \cellcolor{col1}Abs Rel & \cellcolor{col1}Sq Rel & \cellcolor{col1}RMSE  & \cellcolor{col1}RMSE log & \cellcolor{col2}$\delta < 1.25 $ & \cellcolor{col2}$\delta < 1.25^{2}$ & \cellcolor{col2}$\delta < 1.25^{3}$  \\
\hline
~& Monodepth2~\cite{godard2019digging}   &   0.143  &   1.952  &   9.817  &   0.218  &   0.812  &   0.937  &   0.974  \\
~& HR-Depth~\cite{lyu2021hr}   &   0.154  &   2.112  &  10.116  &   0.225  &   0.786  &   0.933  &   0.977  \\
& CADepth~\cite{yan2021channel} &   0.141  &   1.778  &   9.448  &   0.208  &   0.809  &   0.945  &   0.981  \\
Foggy & DIFFNet~\cite{zhou2021self}  &   0.125  &   1.560  &   8.724  &   0.188  &   0.840  &   0.956  &   0.985  \\
& MonoViT~\cite{zhao2022monovit} &   \underline{0.109}  &   \underline{1.206}  &   \underline{7.758}  &   \underline{0.167}  &   \underline{0.870}  &   \underline{0.967}  &   \underline{0.990}  \\
~& \textbf{Robust-Depth} &   0.140  &   1.907  &   9.098  &   0.203  &   0.827  &   0.949  &   0.980  \\
~& \textbf{Robust-Depth$^{\dagger}$} &   0.138  &   1.655  &   9.064  &   0.203  &   0.817  &   0.950  &   0.983  \\
~& \textbf{Robust-Depth$^*$} &   \textbf{0.105}  &   \textbf{1.135}  &   \textbf{7.276}  &   \textbf{0.158} &   \textbf{0.882}  &   \textbf{0.974}  &   \textbf{0.992}  \\
\hline

\hline
\hline
~& Monodepth2~\cite{godard2019digging}  &   0.170  &   2.211  &   8.453  &   0.232  &   0.781  &   0.932  &   0.973  \\
~& HR-Depth~\cite{lyu2021hr}   &   0.173  &   2.424  &   8.592  &   0.237  &   0.783  &   0.927  &   0.972  \\
& CADepth~\cite{yan2021channel} &   0.161  &   2.086  &   8.167  &   0.222  &   0.804  &   0.936  &   0.976  \\
Cloudy & DIFFNet~\cite{zhou2021self}   &   0.154  &   1.839  &   7.679  &   0.212  &   0.809  &   0.941  &   \underline{0.978}  \\
 & MonoViT~\cite{zhao2022monovit} &   \textbf{0.141}  &   \textbf{1.626}  &   \underline{7.550}  &   \textbf{0.201}  &   \textbf{0.831}  &   \textbf{0.948}  &   \textbf{0.981}  \\
~& \textbf{Robust-Depth} &   0.173  &   2.281  &   8.269  &   0.231  &   0.782  &   0.933  &   0.973  \\
~& \textbf{Robust-Depth$^{\dagger}$} &   0.168  &   1.969  &   8.363  &   0.231  &   0.775  &   0.932  &   0.975  \\
~& \textbf{Robust-Depth$^*$} &   \underline{0.148}  &   \underline{1.781}  &   \textbf{7.472}  &   \underline{0.204}  &   \underline{0.825}  &   \underline{0.947}  &   \textbf{0.981}  \\
\hline

\hline
\hline
~& Monodepth2~\cite{godard2019digging}  &   0.245  &   3.641  &  12.282  &   0.310  &   0.600  &   0.852  &   0.945  \\
~& HR-Depth~\cite{lyu2021hr}    &   0.267  &   4.270  &  12.750  &   0.331  &   0.593  &   0.833  &   0.932  \\
& CADepth~\cite{yan2021channel}&   0.226  &   3.338  &  11.828  &   0.288  &   0.633  &   0.874  &   0.956  \\
Rainy & DIFFNet~\cite{zhou2021self} &   0.197  &   2.669  &  10.682  &   0.256  &   0.678  &   0.904  &   0.971  \\
& MonoViT~\cite{zhao2022monovit} &   \underline{0.175}  &   \underline{2.138}  &   \underline{9.616}  &   \underline{0.232}  &   \underline{0.730}  &   \underline{0.931}  &   \underline{0.979}  \\
~& \textbf{Robust-Depth} &   0.199  &   2.670  &  10.595  &   0.260  &   0.677  &   0.902  &   0.972  \\
~& \textbf{Robust-Depth$^{\dagger}$} &   0.182  &   2.250  &  10.241  &   0.247  &   0.711  &   0.910  &   0.973  \\
~& \textbf{Robust-Depth$^*$} &   \textbf{0.167}  &   \textbf{2.019}  &   \textbf{9.157}  &   \textbf{0.221}  &   \textbf{0.755}  &   \textbf{0.938}  &   \textbf{0.982}  \\
\hline

\hline
\hline
~& Monodepth2~\cite{godard2019digging}  &   0.177  &   2.103  &   8.209  &   0.240  &   0.782  &   0.925  &   0.968  \\
~& HR-Depth~\cite{lyu2021hr}    &   0.173  &   1.910  &   7.924  &   0.233  &   0.768  &   0.932  &   0.975  \\
& CADepth~\cite{yan2021channel}&   0.164  &   1.838  &   7.890  &   0.225  &   0.794  &   0.936  &   \underline{0.976}  \\
Sunny & DIFFNet~\cite{zhou2021self}  &   0.162  &   1.755  &   \underline{7.489}  &   0.221  &   0.801  &   0.936  &   0.974  \\
& MonoViT~\cite{zhao2022monovit} &   \textbf{0.150}  &   \underline{1.615}  &   7.657  &   \underline{0.211}  &   \textbf{0.815}  &   \underline{0.943}  &   \textbf{0.979}  \\
~& \textbf{Robust-Depth} &   0.185  &   2.174  &   8.084  &   0.246  &   0.765  &   0.919  &   0.965  \\
~& \textbf{Robust-Depth$^{\dagger}$} &   0.181  &   1.922  &   8.190  &   0.245  &   0.757  &   0.921  &   0.968  \\
~& \textbf{Robust-Depth$^*$} &   \underline{0.152}  &   \textbf{1.574}  &   \textbf{7.293}  &   \textbf{0.210}  &   \underline{0.812}  &   \textbf{0.944}  &   \textbf{0.979}  \\

\hline

\hline

\end{tabular}}
\vspace{1mm}
\caption{\textbf{Results on DrivingStereo Dataset~\cite{yang2019drivingstereo}}. Models were trained on the KITTI dataset and tested on four different domains. Robust-Depth$^*$ represents Robust-Depth trained with MonoViT's backbone and Robust-Depth$^\dagger$ is Robust-Depth trained with weather and time-specific augmentations.}
\label{tab:ds}
\vspace{0.3mm}
\end{table}
We first compare our method using the DrivingStereo dataset to display depth estimation performance in different domains. Specifically, we are testing on foggy, cloudy, rainy and sunny data. We show, in Table \ref{tab:ds}, impressive performance improvements on both the foggy and rainy domains but also difficulties with cloudy and sunny domains. This can be explained by our depth network being to some degree biased towards \textit{bad weather}.
Furthermore, both the foggy and rainy domains demonstrate light weather conditions, and we would expect our results to be more pronounced for more pronounced weather conditions.  
Furthering this point, we show our results on the Foggy CityScape scenario in Table \ref{tab:CSF}, which show significant improvements. As this fog is denser, our models perform better than others. This foggy setting, although synthetic, is more realistic as it represents a more dense foggy layer, and we expect our model to perform better for more extreme weather conditions.

\begin{table}[h!]
\centering
\resizebox{\columnwidth}{!}{%
\renewcommand{\tabcolsep}{1mm}
\begin{tabular}{|l||c|c|c|c|c|c|c|}
\hline 
\multirow{2}{*}{Method} & \multicolumn{7}{c|}{\cellcolor[RGB]{198,224,180}Foggy CityScapes} \\
    &  \cellcolor{col1}Abs Rel & \cellcolor{col1}Sq Rel & \cellcolor{col1}RMSE  & \cellcolor{col1}RMSE log
      & \cellcolor{col2}$\delta < 1.25 $ & \cellcolor{col2}$\delta < 1.25^{2}$ & \cellcolor{col2}$\delta < 1.25^{3}$   \\ \hline
    Monodepth2 \cite{godard2019digging} &   0.208  &   3.093  &  12.447  & 0.337 & 0.656  &   0.842  &   0.917  \\ \hline 
    HR-Depth~\cite{lyu2021hr} &   0.212  &   3.012  &  12.263  & 0.336 & 0.642  &   0.841  &   0.920 \\ \hline
    CADepth~\cite{yan2021channel} &   0.207  &   2.738  &  11.542 & 0.318 &   0.650  &   0.856  &   0.933  \\ \hline 
    DIFFNet$^{\dagger}$~\cite{zhou2021self}  &   0.187  &   2.583  &  11.337 & 0.302 &   0.698  &   0.867  &   0.937  \\ \hline 
    MonoViT~\cite{zhao2022monovit} &   \underline{0.155}  &   1.873  &   9.585 & 0.244   &   \underline{0.771}  &   0.910  &   0.967  \\     \hline
    
    \hline
    \textbf{Robust-Depth} &   0.160  &   1.569  &   \underline{7.912}  &   \underline{0.224}  &   0.757  &   0.937  &   0.982  \\ \hline 
    \textbf{Robust-Depth$^{\dagger}$} &   0.165  &   \underline{1.561}  &   8.073  &   0.227  &   0.750  &   \underline{0.938}  &   \underline{0.983}  \\ \hline
    \textbf{Robust-Depth$^*$} &   \textbf{0.127}  &   \textbf{1.038}  &   \textbf{6.604}  &   \textbf{0.180}  &   \textbf{0.847}  &   \textbf{0.967}  &   \textbf{0.991}  \\ \hline 
    
\end{tabular}%
}
\vspace{1mm}
\caption{\textbf{Results on Foggy CityScape Dataset \cite{SDV18}}. All models are trained on the KITTI dataset.}
\label{tab:CSF}
\vspace{-0.4cm}
\end{table}
For both the Drivingstereo and Foggy CityScape datasets, Robust-Depth trained with MonoViTs backbone returns a similar pattern. When using Mono-ViT as our backbone, we achieve remarkable performance in all domains. Moreover, when Robust-Depth is trained with only weather and time-specific augmentations, we see a respectable performance improvement.  
For a final test of our method, we use the NuScene-Night dataset defined by \cite{wang2021regularizing} to test the capabilities of the model at night. From Table \ref{tab:NuSN}, we find a significant improvement in nighttime understanding from Robust-Depth, and when using MonoViTs backbone the improvements become even more pronounced. It is worth noting that our method outperforms ADDS-DepthNet, a model specifically trained to handle night data.

\begin{table}[h!]
\centering
\resizebox{\columnwidth}{!}{%
\renewcommand{\tabcolsep}{1mm}
\begin{tabular}{|l||c|c|c|c|c|c|c|}
\hline 
\multirow{2}{*}{Method} & \multicolumn{7}{c|}{\cellcolor[RGB]{198,224,180}NuScenes-Night} \\
    &  \cellcolor{col1}Abs Rel & \cellcolor{col1}Sq Rel & \cellcolor{col1}RMSE & \cellcolor{col1}RMSE log
      & \cellcolor{col2}$\delta < 1.25 $ & \cellcolor{col2}$\delta < 1.25^{2}$ & \cellcolor{col2}$\delta < 1.25^{3}$   \\ \hline
    Monodepth2 \cite{godard2019digging}  &   0.397  &   6.206  &  14.569 & 0.568 &  0.378  &   0.650  &   0.794  \\ \hline
    HR-Depth \cite{lyu2021hr}  &   0.461  &   6.633  &  15.028  &   0.622  &   0.301  &   0.571  &   0.749  \\ \hline
    CADepth \cite{yan2021channel}  &   0.421  &   5.949  &  14.509 &   0.593 &   0.331  &   0.613  &   0.776  \\ \hline
    DIFFNet \cite{zhou2021self}  &   0.344  &   4.851  &  13.152 &   0.491 &   0.440  &   0.710  &   0.838  \\\hline
    ADDS-DepthNet \cite{liu2021self} &   0.415  &   5.693  &  14.905  &   0.590  &   0.343  &   0.594  &   0.766  \\ \hline
    MonoViT \cite{zhao2022monovit}  &   \underline{0.313}  &   \underline{4.143}  &  \underline{12.252}  &   \underline{0.455}  &   0.485  &   0.736  &   0.858  \\         
    \hline
    
    \hline
    \textbf{Robust-Depth}&   0.355  &   6.344  &  12.510  &   0.457  &   \underline{0.523}  &   \underline{0.762}  &   \underline{0.862}  \\ \hline
    \textbf{Robust-Depth$^{\dagger}$} &   0.344  &   4.873  &  12.290  &   0.458  &   0.461  &   0.752  &   \underline{0.862}  \\ \hline
    \textbf{Robust-Depth$^*$} &   \textbf{0.276}  &   \textbf{4.075}  &  \textbf{10.470}  &   \textbf{0.380}  &   \textbf{0.607}  &   \textbf{0.819}  &   \textbf{0.912}  \\
    \hline

\end{tabular}%
}
\vspace{1mm}
\caption{\textbf{Results on NuScenes-Night data \cite{caesar2020nuscenes}}. All models were trained at a resolution of 640$\times$192 except ADDS-DepthNet at 256$\times$512. ADDS-DepthNet is trained with RobotCar dataset \cite{RobotCarDatasetIJRR}.}
\label{tab:NuSN}
\vspace{-0.5cm}
\end{table}
\section{Conclusion}
In this paper, we show how self-supervised MDE can overcome adverse weather effects, a major obstacle to real-world applications (e.g. autonomous vehicles). 
The solution we put forward is a carefully constructed, architecture-agnostic data-augmentation scheme. Central to our scheme is bi-directional pseudo-supervision loss, a novel loss function which uses unaugmented depth estimations to self-supervise the augmented depth estimations and vice versa. The paper also introduces several other algorithmic steps, each of which improves depth estimation even under extremely harsh image degradations. This is done without significant increases in memory requirements or computing. The set of data augmentations we use enables the depth network to extract more varied and reliable cues for depth, leading to significantly better depth estimation than SotA in adverse weather, and equally good or better than SotA even in fine weather. The same proposed techniques can be used in other pixel-wise estimation tasks in the future to improve their robustness to similar augmentations.


{\small
\bibliographystyle{ieee_fullname}
\bibliography{egpaper_final}
}

\clearpage

\section{Supplementary Material}

\subsection{Ablation}

To understand the effects of the augmentations chosen, we carry out another ablation study, first breaking down the augmentations into three categories; weather, image degradation and positional augmentation, where weather also contains time-related augmentations. 
We use Robust-Depth to train individual models, selecting just the augmentations from each category. The results of each of these models, when tested on both the \textit{sunny} and \textit{bad weather} data, are shown in Table \ref{tab:ablation1}. Robust-Depth uses a CNN-based architecture throughout the experiments in Tables \ref{tab:ablation1} and \ref{tab:ablation2}. An interesting finding of this study is that positional augmentations seem to significantly improve the capability of the depth network on unaugmented images. This signifies that the positional augmentations are helping the network develop a wider variety of cues to estimate depth.

\begin{table}[h!]
\centering
\resizebox{\columnwidth}{!}{%

\renewcommand{\tabcolsep}{2mm}
\begin{tabular}{|l|l||c|c|c|c|c|c|c|}  \hline 
    Method & Tests & \cellcolor{col1}Abs Rel & \cellcolor{col1}Sq Rel & \cellcolor{col1}RMSE 
    & \cellcolor{col2}$\delta < 1.25 $ & \cellcolor{col2}$\delta < 1.25^{2}$ & \cellcolor{col2}$\delta < 1.25^{3}$   \\ 
    \hline
    
    \multirow{2}{*}{{Robust-Depth}}  & \textit{Sunny} &   0.115  &   0.937  &   4.840   &   0.873  &   0.959  &   0.981  \\
    & \cellcolor{lightgray} \textit{Bad w.} & \cellcolor{lightgray}  \textbf{0.133}  & \cellcolor{lightgray} {1.115}  &  \cellcolor{lightgray} \textbf{5.259}   & \cellcolor{lightgray}  \textbf{0.842}  &  \cellcolor{lightgray} \textbf{0.948}  &  \cellcolor{lightgray} \textbf{0.977}  \\ 
    \hline
    
    \multirow{2}{*}{{Weather}} & \textit{Sunny} &   0.120  &   \textbf{0.889}  &   4.845    &   0.864  &   0.958  &   0.981  \\
    & \cellcolor{lightgray} \textit{Bad w.} & \cellcolor{lightgray}  0.145  &  \cellcolor{lightgray} \textbf{1.089}  &  \cellcolor{lightgray} 5.512  & \cellcolor{lightgray}  0.808  &  \cellcolor{lightgray} 0.935  &  \cellcolor{lightgray} 0.974  \\ 
    \hline
    
    \multirow{2}{*}{{Img. degradation}} & \textit{Sunny} &   0.123  &   1.000  &   5.049  &   0.860  &   0.954  &   0.979  \\
    & \cellcolor{lightgray} \textit{Bad w.} & \cellcolor{lightgray}  0.181  & \cellcolor{lightgray}  1.654  &  \cellcolor{lightgray} 6.512   & \cellcolor{lightgray}  0.741  &  \cellcolor{lightgray} 0.900  & \cellcolor{lightgray}  0.953  \\
    \hline
    
    \multirow{2}{*}{{Positional aug.}} & \textit{Sunny} &   \textbf{0.111}  &   {0.897}  &   \textbf{4.740}   &   \textbf{0.884}  &   \textbf{0.961}  &   \textbf{0.982}  \\
    & \cellcolor{lightgray} \textit{Bad w.} &  \cellcolor{lightgray} 0.301  &  \cellcolor{lightgray} 3.002  & \cellcolor{lightgray}  9.268   &  \cellcolor{lightgray} 0.510  & \cellcolor{lightgray}  0.760  & \cellcolor{lightgray}  0.878  \\ 
    \hline 
\end{tabular}%
}
\caption{\textbf{Ablation 2:} We split the augmentations into three categories; weather, corruption and positional augmentations. Each uses pretrained ImageNet \cite{deng2009imagenet} weights and a data resolution of 640$\times$192.}
\label{tab:ablation1}
\vspace{-0.4cm}
\end{table}

As would be expected, positional augmentations does not help the network with other domains or lead to greater overall robustness. Furthermore, the \textit{bad weather} test, which contains weather and image degradation augmentations, sees the greatest benefit when training with all augmentations. Indicating that no single augmentation is most beneficial for multiple domains.

\begin{table}[h!]
\centering
\resizebox{\columnwidth}{!}{%

\renewcommand{\tabcolsep}{2mm}
\begin{tabular}{|l|l||c|c|c|c|c|c|c|}
\hline 
    Method & Tests & \cellcolor{col1}Abs Rel & \cellcolor{col1}Sq Rel & \cellcolor{col1}RMSE 
      & \cellcolor{col2}$\delta < 1.25 $ & \cellcolor{col2}$\delta < 1.25^{2}$ & \cellcolor{col2}$\delta < 1.25^{3}$   \\ \hline

    \multirow{2}{*}{\textbf{Vertical}} & \textit{Sunny} &   0.120  &   0.995  &   4.949    &   0.868  &   0.957  &   0.980  \\
    & \cellcolor{lightgray} \textit{Bad w.} &  \cellcolor{lightgray} 0.288  & \cellcolor{lightgray}  3.194  & \cellcolor{lightgray}  8.597  &  \cellcolor{lightgray} 0.555  &  \cellcolor{lightgray} 0.790  &  \cellcolor{lightgray} 0.901  \\ \hline

    \multirow{2}{*}{\textbf{Tile}} & \textit{Sunny} &   \textbf{0.117}  &   \textbf{0.901}  &   \textbf{4.819}   &   \textbf{0.871}  &   \textbf{0.958}  &   \textbf{0.981}  \\
    & \cellcolor{lightgray} \textit{Bad w.} &  \cellcolor{lightgray} 0.313  &  \cellcolor{lightgray} 3.184  &  \cellcolor{lightgray} 9.475  & \cellcolor{lightgray}  0.497  &  \cellcolor{lightgray} 0.748  &  \cellcolor{lightgray} 0.875  \\ \hline

    \multirow{2}{*}{\textbf{Rand. Erase}} & \textit{Sunny} &   0.119  &   0.985  &   4.953  &   \textbf{0.871}  &   0.957  &   0.980  \\
    & \cellcolor{lightgray} \textit{Bad w.} & \cellcolor{lightgray}  \textbf{0.256}  &  \cellcolor{lightgray} \textbf{2.368}  &  \cellcolor{lightgray} \textbf{7.967}  &   
    \cellcolor{lightgray} \textbf{0.589}  &  \cellcolor{lightgray} \textbf{0.819}  &  \cellcolor{lightgray} \textbf{0.921}  \\ \hline

    \multirow{2}{*}{\textbf{Scale}} & \textit{Sunny} &   0.119  &   1.040  &   4.937  &   0.869  &  \textbf{0.958}  &   \textbf{0.981}  \\
    & \cellcolor{lightgray} \textit{Bad w.} &  \cellcolor{lightgray} 0.300  & \cellcolor{lightgray}  3.022  & \cellcolor{lightgray}  8.927   &  \cellcolor{lightgray} 0.525  &  \cellcolor{lightgray} 0.769  &  \cellcolor{lightgray} 0.884  \\ \hline

\end{tabular}%
}
\caption{\textbf{Ablation 3:} We further break down positional augmentations into vertical cropping, tiling cropping, random erase and scaling. Each uses pretrained ImageNet \cite{deng2009imagenet} weights and a data resolution of 640$\times$192.}
\label{tab:ablation2}
\vspace{-0.4cm}
\end{table}

We now further explore each subcategory. We break positional augmentations into its components; vertical cropping, tiling, random erase and scaling. Table \ref{tab:ablation2} demonstrates that each individual positional augmentation does not lead to an improved performance for depth estimation when compared to the baseline of Monodepth2 \cite{godard2019digging}. We believe these positional augmentations largely benefit from each other, and an over-reliance on each individual augmentation leads to the worsening of the depth network's standard cues. 

Another interesting finding shown in Table \ref{tab:ablation2}, is that tile cropping augmentation gives rise to the lowest \textit{sunny} error, suggesting that a greater local region understanding is the most beneficial feature of positional augmentation, at least for a CNN-based backbone. On top of that, random erase leads to the best robust performance for \textit{bad weather} testing. This is because random erase aims to improve the model's capabilities with occlusion, and many weather and corruption-related augmentations would benefit from this. 


\subsection{Eigen Benchmark}

We provide the test results from the KITTI dataset with the improved ground truth data. The improved ground truth results, shown in Table \ref{tab:eigentest}, look very similar to Table 2 from the main paper. Robust-Depth can maintain \textit{sunny} depth quality while improving the quality on the \textit{bad weather} test set. In other words, it is more robust to weather changes and image degradation while maintaining capabilities in sunny scenes. Furthermore, Robust-Depth$^*$, uses MonoViT \cite{zhao2022monovit} as a backbone and shows greater overall performance for \textit{bad weather} testing yet competitive performance for the \textit{sunny} test.

\begin{table}[h!]
\centering
\resizebox{\columnwidth}{!}{%

\renewcommand{\tabcolsep}{2mm}
\begin{tabular}{|l|l||c|c|c|c|c|c|c|}
\hline 
    Method & Tests & \cellcolor{col1}Abs Rel & \cellcolor{col1}Sq Rel & \cellcolor{col1}RMSE 
      & \cellcolor{col2}$\delta < 1.25 $ & \cellcolor{col2}$\delta < 1.25^{2}$ & \cellcolor{col2}$\delta < 1.25^{3}$   \\ \hline

        \multirow{2}{*}{Monodepth2 \cite{godard2019digging}} & \textit{Sunny} & 0.090 & 0.545 & 3.942  & 0.914 & 0.983 & 0.995 \\ 
         &  \cellcolor{lightgray} \textit{Bad w.} &  \cellcolor{lightgray} 0.223  &  \cellcolor{lightgray} 2.136  &  \cellcolor{lightgray} 7.464    & \cellcolor{lightgray}  0.654  & \cellcolor{lightgray}  0.850  & \cellcolor{lightgray}  0.931   \\
         \hline
         
        \multirow{2}{*}{HR-Depth~\cite{lyu2021hr}}  & \textit{Sunny} & 0.085 & 0.471 & 3.769  & 0.919 & 0.985 & 0.996\\
        & \cellcolor{lightgray} \textit{Bad w.} & \cellcolor{lightgray}  0.251  & \cellcolor{lightgray}  2.331  &  \cellcolor{lightgray} 8.093   &  \cellcolor{lightgray} 0.590  &  \cellcolor{lightgray} 0.814  & \cellcolor{lightgray}  0.912  \\
        \hline
         
        \multirow{2}{*}{CADepth~\cite{yan2021channel}}  & \textit{Sunny} & 0.080 & 0.450 & 3.649  & 0.927 & 0.986 & 0.996\\ 
        & \cellcolor{lightgray}  \textit{Bad w.} & \cellcolor{lightgray}  0.243  & \cellcolor{lightgray}  2.252  &  \cellcolor{lightgray} 7.761    & \cellcolor{lightgray}  0.611  & \cellcolor{lightgray}  0.824  &  \cellcolor{lightgray} 0.919  \\
        \hline
         
        \multirow{2}{*}{DIFFNet$^{\dagger}$~\cite{zhou2021self}}  & \textit{Sunny} & 0.076 & 0.412 & 3.494  & 0.935 & 0.988 & 0.996\\
        & \cellcolor{lightgray} \textit{Bad w.} & \cellcolor{lightgray}  0.183  & \cellcolor{lightgray}  1.542  & \cellcolor{lightgray}  6.842    &  \cellcolor{lightgray} 0.717  &  \cellcolor{lightgray} 0.888  &  \cellcolor{lightgray} 0.949  \\
        \hline
        
        \multirow{2}{*}{MonoViT~\cite{zhao2022monovit}}  & \textit{Sunny} & \textbf{0.075} & \textbf{0.389} & \textbf{3.419}  & \textbf{0.938} & \textbf{0.989} & \textbf{0.997}\\
        & \cellcolor{lightgray} \textit{Bad w.} & \cellcolor{lightgray}  0.148  &  \cellcolor{lightgray} 1.133  &  \cellcolor{lightgray} 5.931   & \cellcolor{lightgray}  0.785  & \cellcolor{lightgray}  0.930  & \cellcolor{lightgray}  0.972  \\
        \hline
        
        \hline
        \multirow{2}{*}{\textbf{Robust-Depth}}  & \textit{Sunny} &   0.091  &   0.579  &   3.975    &   0.912  &   0.981  &   0.994  \\
        & \cellcolor{lightgray} \textit{Bad w.} &  \cellcolor{lightgray} 0.110  & \cellcolor{lightgray}  0.777  & \cellcolor{lightgray}  4.511   &  \cellcolor{lightgray} 0.879  &  \cellcolor{lightgray} 0.969  &  \cellcolor{lightgray} 0.990  \\
        \hline
        
        \multirow{2}{*}{\textbf{Robust-Depth$^*$}}  & \textit{Sunny} &   0.077  &   0.417  &   3.548    &   0.932  &   0.988  &   \textbf{0.997} \\
        & \cellcolor{lightgray} \textit{Bad w.} &  \cellcolor{lightgray} \textbf{0.093}  & \cellcolor{lightgray}  \textbf{0.583}  &  \cellcolor{lightgray} \textbf{4.130}   &   \cellcolor{lightgray} \textbf{0.904}  &  \cellcolor{lightgray} \textbf{0.979}  &  \cellcolor{lightgray} \textbf{0.994}  \\ \hline
\end{tabular}}
\caption{\textbf{Eigen improved ground truth test:} All tests are performed at a resolution of 640$\times$192 and pretrained with ImageNet \cite{deng2009imagenet} weights.}
\label{tab:eigentest}
\vspace{-0.4cm}
\end{table}

\begin{figure*}[h!]
\begin{center}
  \includegraphics[width=1\linewidth]{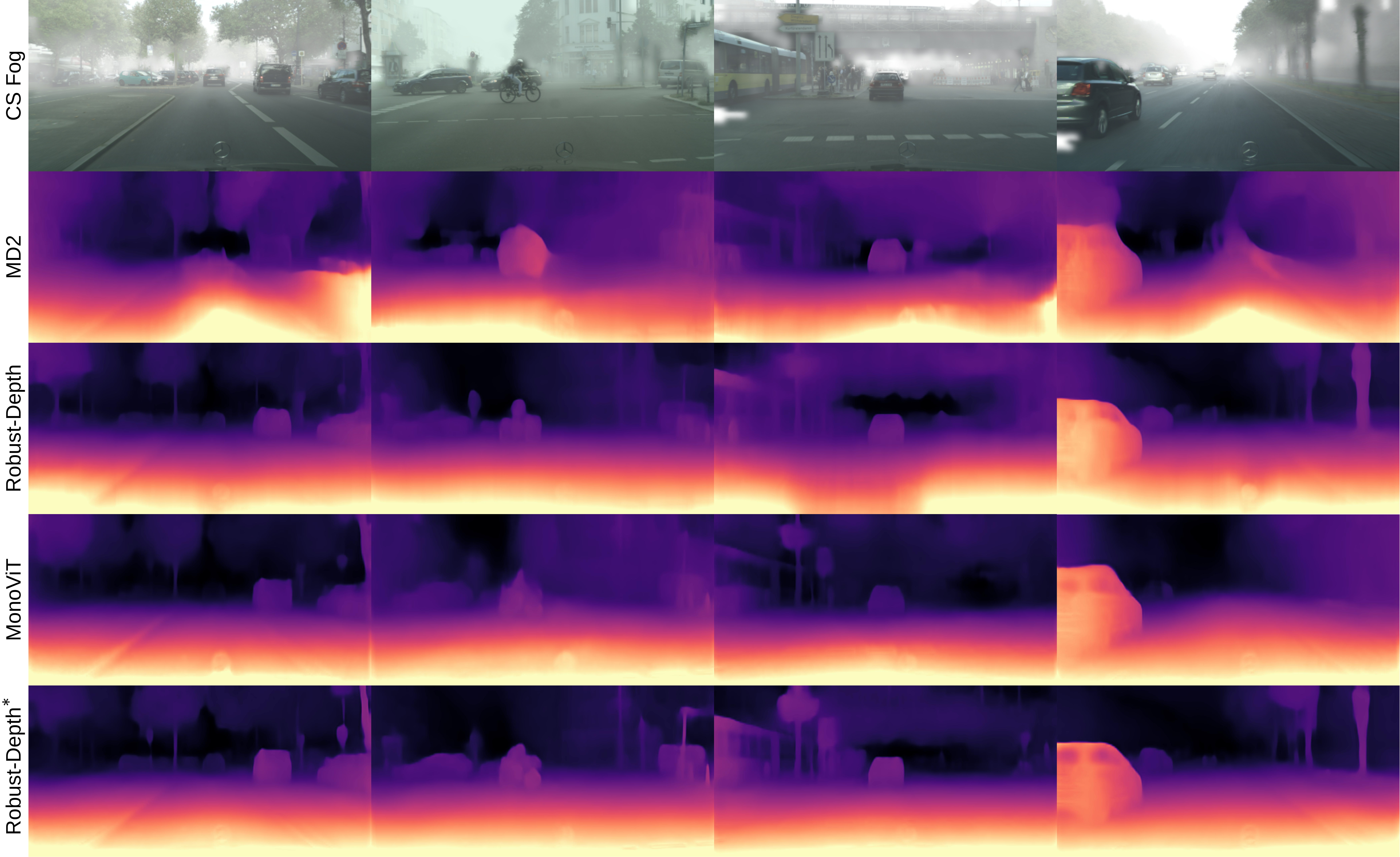}
\end{center}
  \caption{Demonstrating the qualitative results on the Foggy CityScape test dataset. } 
\label{CSFOG}
\end{figure*}

\subsection{Qualitative results}

To show how our model can handle changes in domains, we also present qualitative results from some out-of-distribution data. Specifically, we will be looking at DrivingStereo \cite{yang2019drivingstereo}, Foggy CityScape \cite{SDV18, cordts2016cityscapes} and Nuscenes-Night \cite{caesar2020nuscenes}.
 
\begin{figure*}[h]
\begin{center}
  \includegraphics[width=1\linewidth]{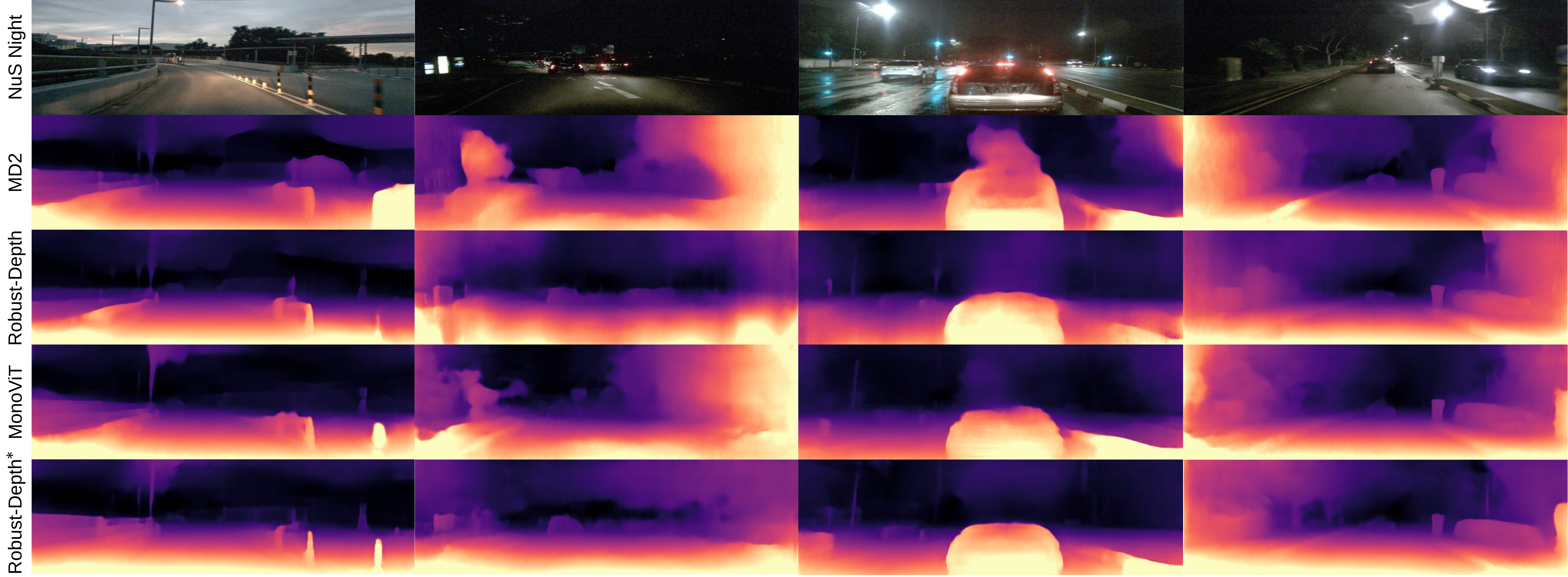}
\end{center}
  \caption{Demonstrating the qualitative results on the NuScenes Night test dataset.} 
\label{NUNight}
\end{figure*}

Figure \ref{CSFOG} clearly demonstrates the visual improvements in our method compared with Monodepth2 and MonoViT. Our method learns to ignore fog in scenes and predict realistic depth. Methods like Monodepth2 display poor depth estimations in foggy scenes, and even current SotA models are unable to reconstruct sharp edges around objects when in the foggy domain. Robust-Depth generalises to this dataset and solves both issues without seeing this dataset.  

In Figure \ref{NUNight}, we evaluate our method in the nighttime domain. We see this is a much more challenging domain due to illuminations and lack of texture changes. Nevertheless, results indicate that our model can more clearly see objects and infer smoother surfaces. 

\begin{figure*}[h]
\begin{center}
  \includegraphics[width=1\linewidth]{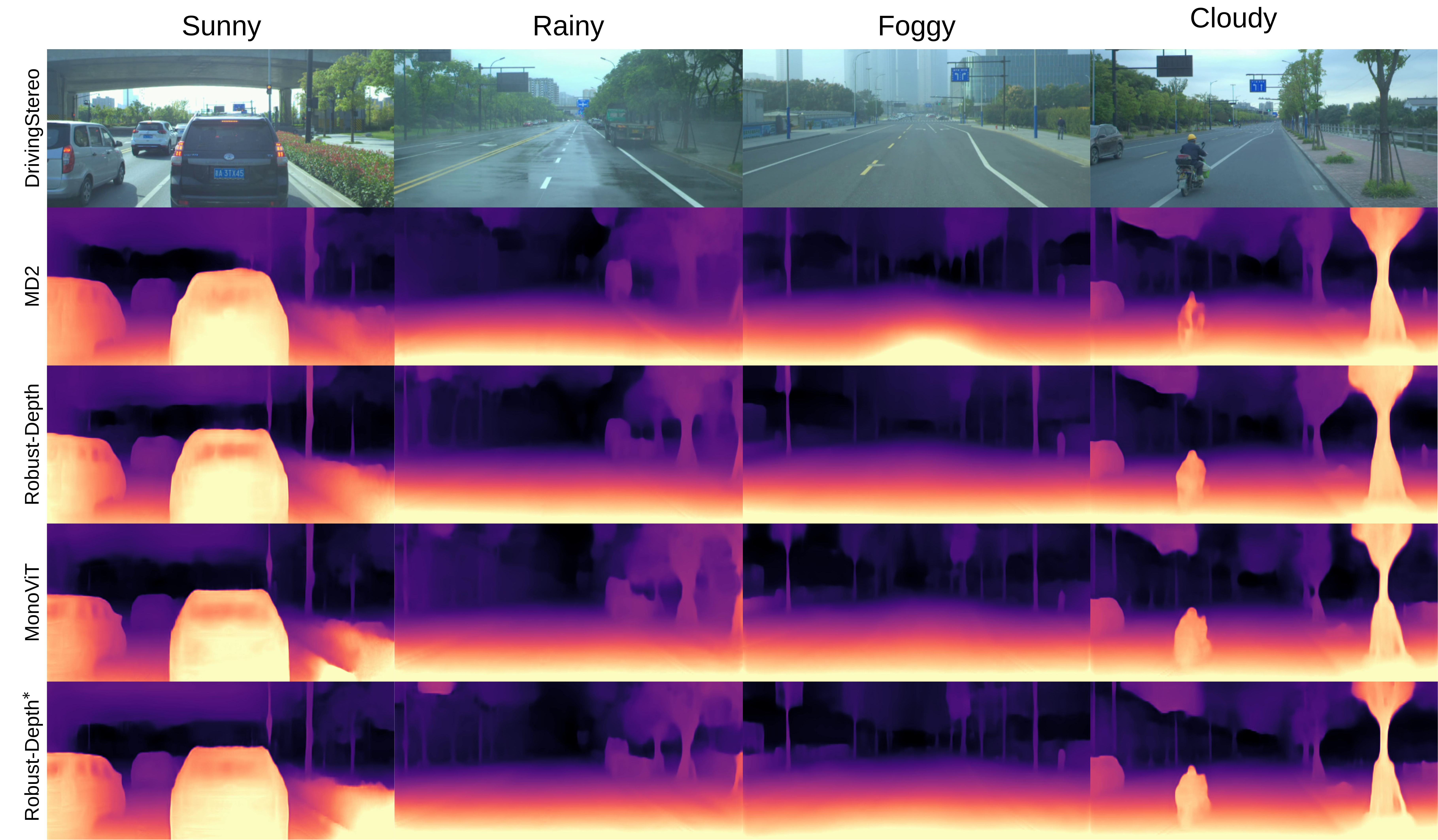}
\end{center}
  \caption{A demonstration of the qualitative results from the DrivingStereo test dataset.} 
\label{Drivingstereo}
\end{figure*}

Furthermore, we look at the DrivingStereo dataset with all four domains; sunny, rainy, foggy and cloudy in Figure \ref{Drivingstereo}. Clear and significant improvements can be seen when comparing Robust-Depth to Monodepth2. Also, the difference between MonoViT and our Robust-Depth$^*$ shows finer advancements in all presented qualitative results. Most improvements with this backbone involve finer details in edge definition and smoother depth maps. 

 Further qualitative results on the KITTI Eigen test are shown in Figures \ref{1st} and \ref{2nd}. These figures show the collection of augmentations used during the training of our method. Here we notice significant improvements from each augmentation, especially when looking at rain-related images, compared to Monodepth2.
 Interestingly, Robust-Depth removes dependence on colour channels and colour overall. Also, when looking at Gaussian/impulse/shot noise, we witness remarkable improvements when using both backbones from our method. 
 
 We can also explore the understanding of these depth networks in how they hallucinate depth from a single image based on our results. We observe that:
 \begin{itemize}
  \item Colour channels are not vital in determining depth (Figure \ref{2nd} row four)
  \item Robust-Depth does not need to rely on vertical cues (Figure \ref{1st} row one, column four)
  \item Robust-Depth can handle occlusions (Figure \ref{1st} rows one, column three)
  \item Robust-Depth can understand texture changes (Figure \ref{1st} rows two-five)
\end{itemize}
We show that we can infer depth from a very wide variety of images and not significantly negatively affect depth performance. Giving evidence that our model uses a much wider assortment of cues for monocular depth estimation.

\subsection{Bi-directional pseudo-supervised depth loss:}
\begin{figure}[h!]
  \centering
   \includegraphics[width=1\linewidth]{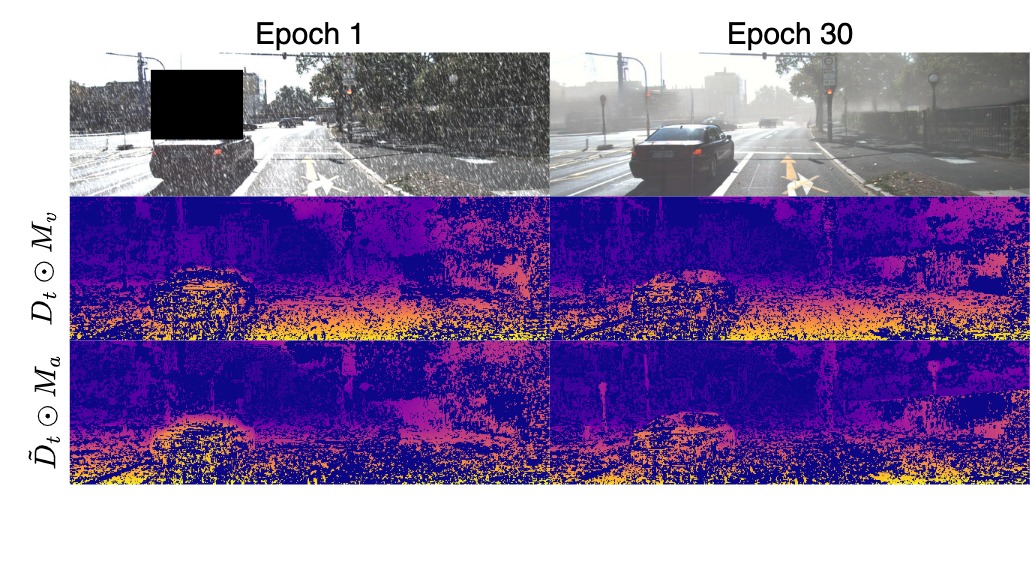}
   \caption{Depth masks $D_t \odot M_v$ and $\tilde{D}_t \odot M_a$ are both shown for epoch one and epoch thirty out of thirty epochs.}
   \label{fig:masks}
\end{figure}

The pseudo-supervised depth loss, as discussed in the main text, allows our augmented and unaugmented depths to pseudo-supervise each other. In this section, we will discuss the effects of each depth estimation on the final loss. In the beginning stages of learning the model capitalises on the use of two depth maps to learn depth faster, as augmented images are a view of the same image with variations in texture, shading and illumination patterns. The depth maps teach each other and result in faster learning. In Figure \ref{fig:masks}, we visualise the masks described in equations 8 and 9 from the main text, multiplied by the depth. 

For the first column, $D_t \odot M_v$ is the unaugmented depth pixels that result in the lowest reprojection error. On the other hand, $\tilde{D}_t \odot M_a$ is the augmented depth pixels that result in the lowest reprojection error. We see throughout training that the unaugmented depth is moderately more accurate than the augmented depth estimation. This suggests that unaugmented depth will have a larger weight to the bi-directional depth loss throughout training. However, augmented depth will still hold a significant influence as many pixels of the augmented depth estimation lead to greater reprojections (row three), encouraging the use of a bi-directional depth loss.


\subsection{Practicalities of vertical cropping and tiling}\label{vert+tile}

\begin{figure}[h!]
  \centering
   \includegraphics[width=1\linewidth]{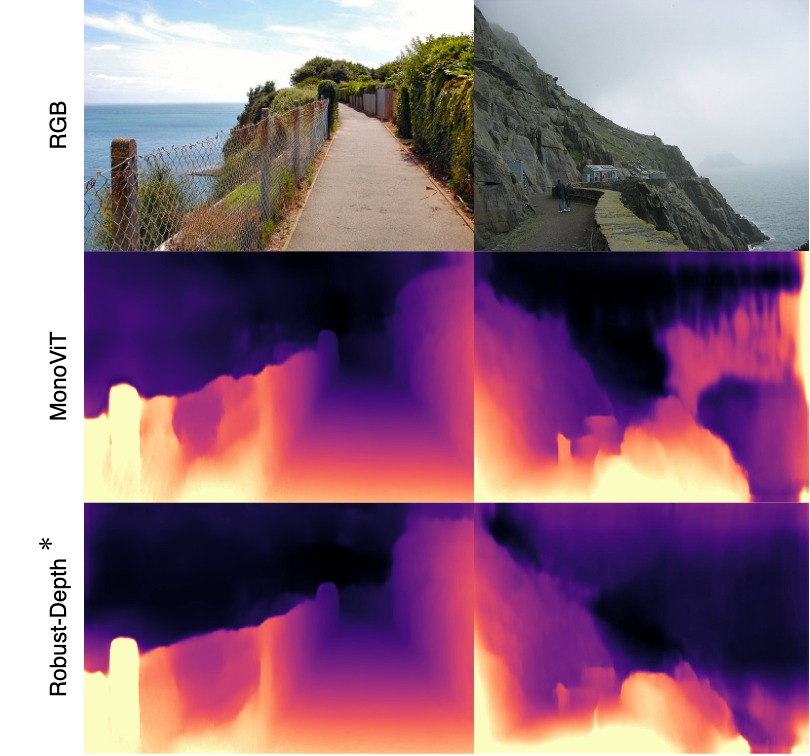}
   \caption{Image from column one \cite{clif2} and image from column two \cite{clif1} demonstrates that Robust-Depth$^*$ is more likely to assume large gradient changes than previous methods}
   \label{fig:vertical}
\end{figure}

As discussed in the introduction, vertical and tile cropping help the depth network to remove pixel positional dependencies and learn that the lower sections of an image are not always close to the camera. In Figure \ref{fig:vertical}, we see two examples of cliff-side edges, in both scenarios, it would be dangerous to assume that the pixels closest to the ground represent the road.
When using our vertical/tile cropping we can observe improvements in the understanding of depth. As the network is not over-reliant on vertical cues, it can assume that there are large gradient changes over walls. 

\subsection{Limitations}
\begin{figure}[h!]
  \centering
   \includegraphics[width=0.85\linewidth]{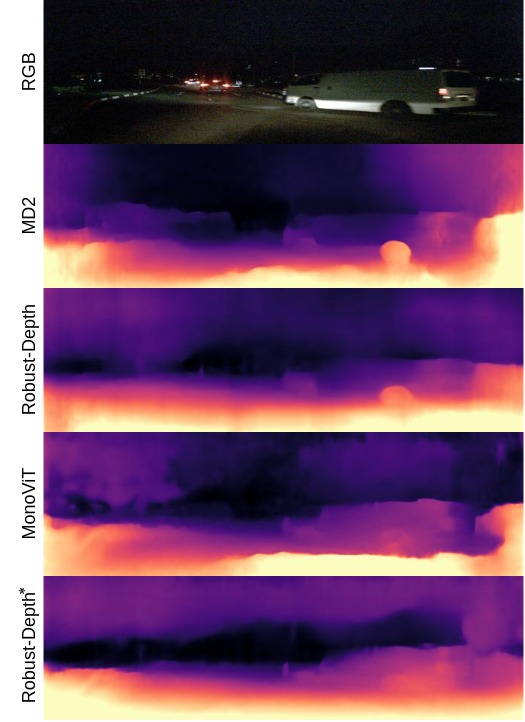}
   \caption{Vehicles disappearing in dark scenes.}
   \label{fig:nightdiss}
\end{figure}

There are still many limitations to self-supervised monocular depth estimation. 
As we know from \cite{dijk2019neural} self-supervised monocular depth relies on many naive cues, specifically when looking at Figure \ref{fig:nightdiss}, we see that in nighttime scenes, our method, as well as the current state-of-the-art methods, cannot accurately detect vehicles. We believe this is because these methods use the shadows under the vehicles to determine the object's depth \cite{dijk2019neural}, and with nighttime scenes, this cue cannot be relied upon. When we use CoMoGAN \cite{pizzati2021comogan} to generate nighttime augmentations, we see that, although realistic, the night scene maintains some shadow structures underneath the cars. To improve upon this flaw in the future, the method of augmentation chosen can focus on recreating even harsher nighttime scenes, removing any indication of shadows underneath the vehicle. 

Moreover, even with the use of vertical cropping and tiling (section \ref{vert+tile}), there is still a lack of understanding of large distance change (see Figure \ref{fig:vertical}). This is an area for future work.

\begin{figure}[h!]
  \centering
   \includegraphics[width=0.85\linewidth]{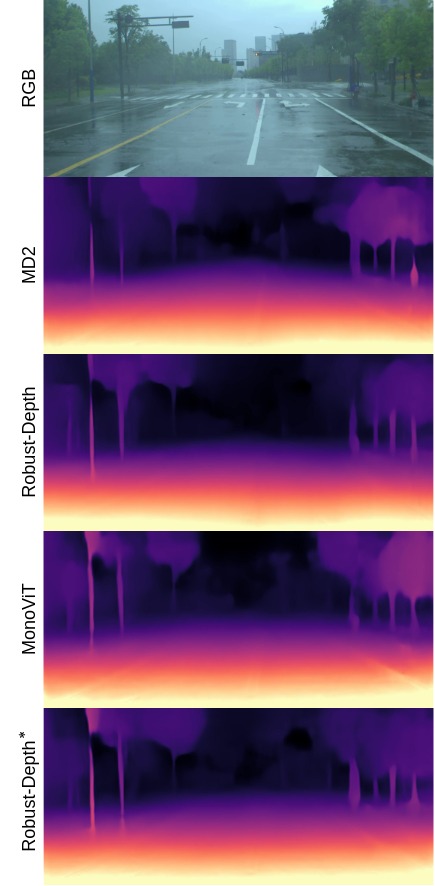}
   \caption{Trees forming out of lampposts and reflections causing errors. The depth map is generated using Robust-Depth$^*$.}
   \label{fig:lamp}
\end{figure}

Furthermore, due to the over-reliance on the KITTI dataset, we overfit on tree structures. Figure \ref{fig:lamp} shows that the lampposts are being reconstructed as trees because the KITTI dataset has many examples of trees and fewer examples of tall lampposts. A simple solution to this problem is to train on a greater variety of data. 

From Figure \ref{fig:lamp}, we can also see that there are reflections caused by the rain, leading to the depth network inferring depth on the reflected lamppost. This, although a simple-looking problem, requires a sophisticated understanding of the scene and reflections themselves. A potential solution from an augmentation perspective is to use more realistic GANs that facilitate the creation of wet scenes that contain many reflections. 
\subsection{Data creation}

As discussed in the main text, we generate any computationally expensive augmentations before training, which speeds up the training process. However, depending on the augmentations chosen, our method could be trained end-to-end. For example, vertical cropping, tiling, random erase, colour jitter, horizontal flips and scaling are all randomised and applied during training. 
On the other hand, we create dusk, dawn and night version of the KITTI dataset using CoMoGANs pretrained model. Furthermore, we create a realistic rainy version of the KITTI dataset using a physics-based render \cite{tremblay2021rain} and a GAN trained on the NuScenes rainy data. The GAN \cite{CycleGAN2017} converts the KITTI images from clear to rainy, creating reflections, rain on the camera, and creating desaturated scenes. 
Then, we apply the physics-based render, where we specify a volume of rainfall per KITTI scene which will be described on the project's GitHub page.
Also, using the physics base renderer, we create foggy scenes, which have parameters of beta set to random for training and set to 1 for all test images. 

\begin{figure*}[h!]
\begin{center}
  \includegraphics[width=0.95\linewidth]{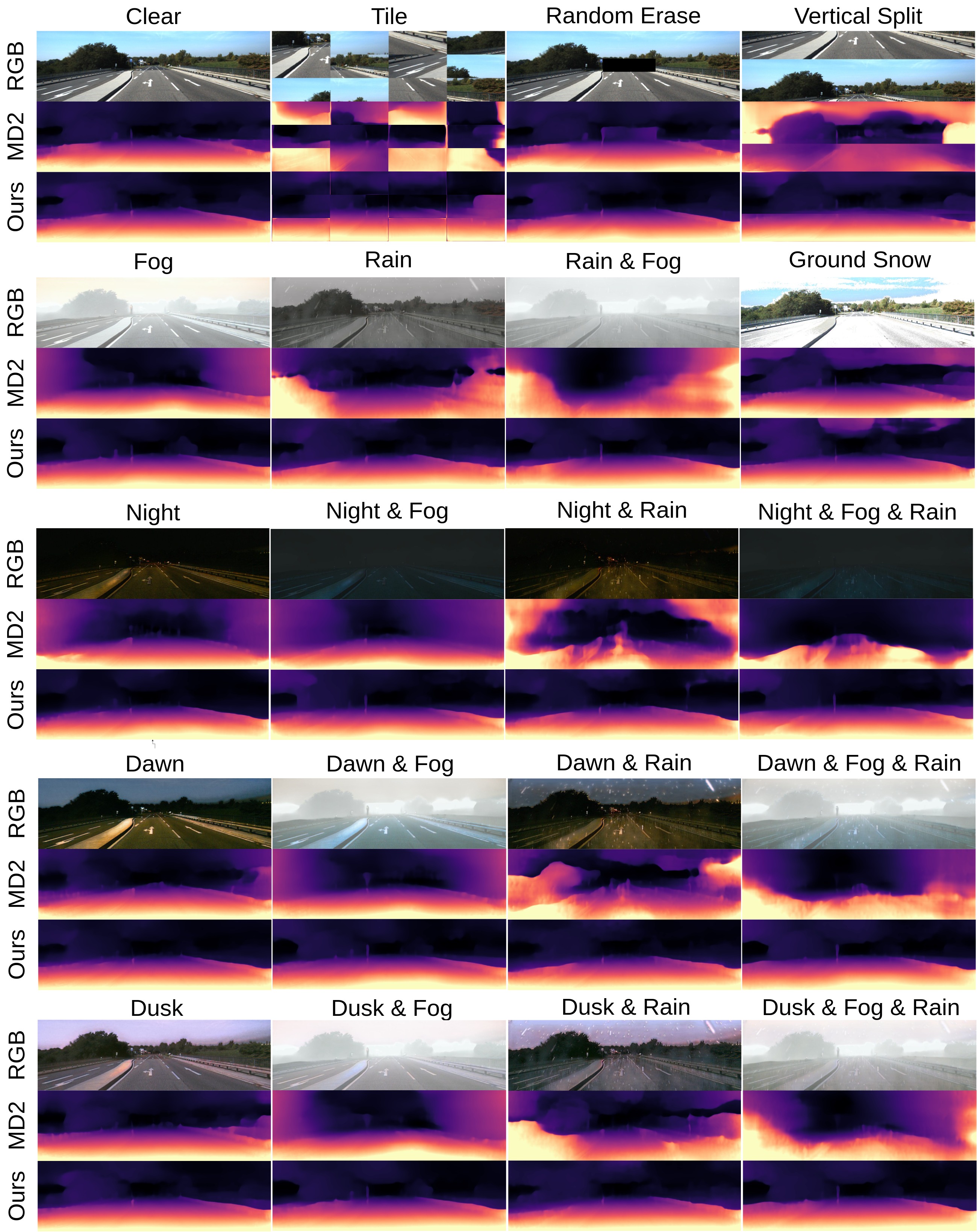}
\end{center}
  \caption{We demonstrate a majority of the weather-related and positional augmentations. MD2 represents depth estimations using Monodepth2, and "Ours" is Robust-Depth. All images are from the KITTI Eigen test data.} 
\label{1st}
\end{figure*}

\begin{figure*}[h!]
\begin{center}
  \includegraphics[width=0.95\linewidth]{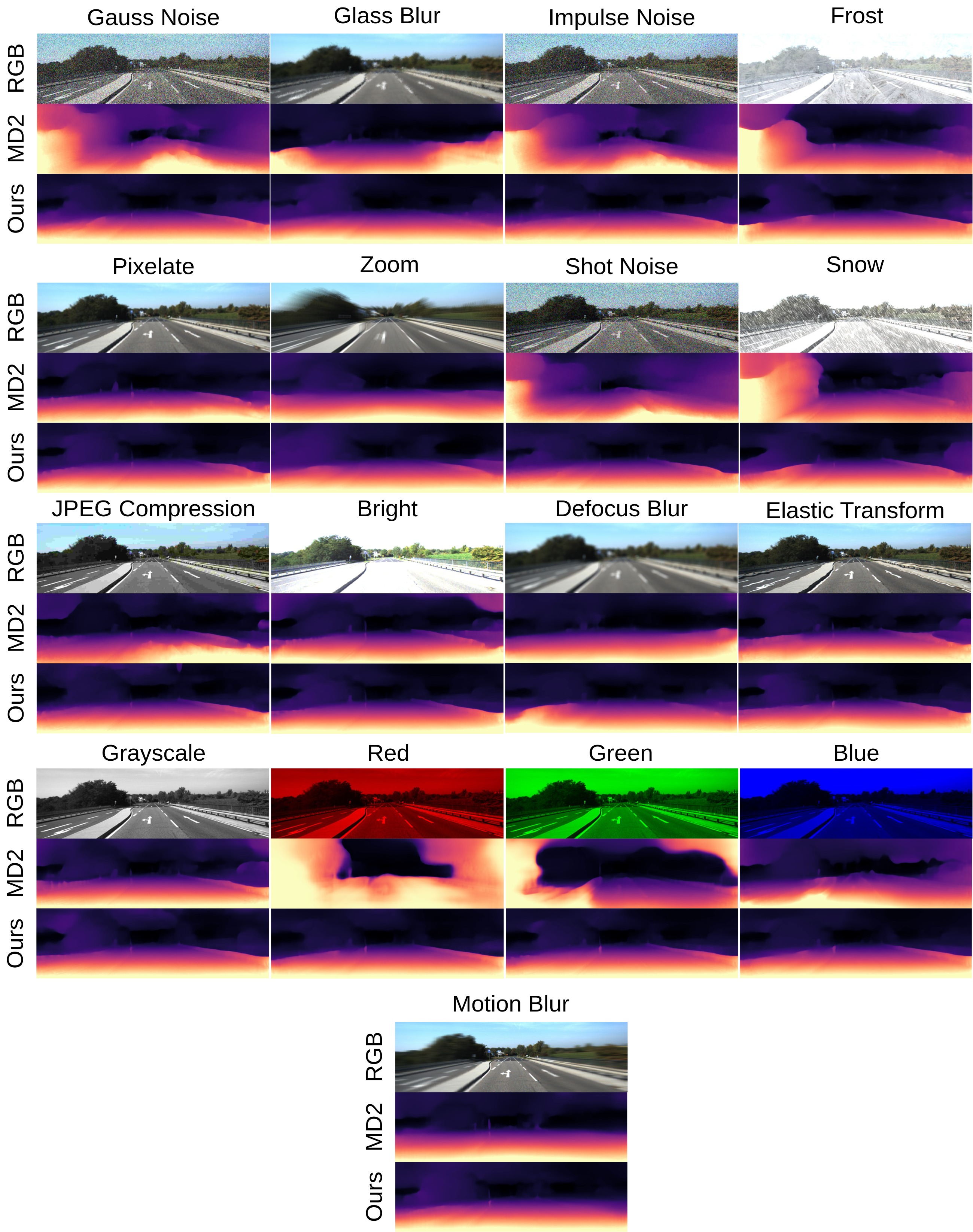}
\end{center}
  \caption{Corruptions, colour channels, greyscale and motion blur augmentations are shown. MD2 represents depth estimations using Monodepth2, and "Ours" is Robust-Depth. All images are from the KITTI Eigen test data.} 
\label{2nd}
\end{figure*}

At this point, we create all combinations of the augmented data, as follows: Dusk, Dawn, Night, Rain, Fog, Rain+Fog, Rain+Dawn, Rain+Night, Rain+Dusk, Fog+Night, Fog+Dusk, Fog+Dawn, Rain+Fog+Night, Rain+Fog+Dawn and Rain+Fog+Dusk. We also add motion blur as it can negatively affect self-supervised depth estimation, as well as ground snow to represent more extreme weather. Both of these augmentations were created using the Automold GitHub page \cite{Automold}, and the severity of the augmentations were set to random for training but max severity for the test data. Note that the \textit{Bad weather} test contains augmentations from weather, time of day and image degradation, but no positional augmentations. 
To create the corrupted data, we directly use the code provided by \cite{hendrycks2019robustness}. Here we set the severity of each corrupted image to the maximum for testing data, but random for the training data. Finally, we create simple greyscale, red, green and blue components of the images. All of these augmentations are set to have a uniform distribution of being selected, without replacement, so each augmentation is sampled equally during training. The augmented data represents half of the data seen during training and each augmentation has a 1/$n$ chance of being selected, where $n$ is the number of augmentations chosen. All the information provided should aid with the reproducibility of this work and potential further development. We highlighted in Figure \ref{1st} and Figure \ref{2nd} a multitude of cases, that clearly demonstrate the improvements of our model (Robust-Depth) over Monodepth2 (MD2). 

\end{document}